\documentclass[journal]{IEEEtran}

\ifCLASSINFOpdf
  \usepackage[pdftex]{graphicx}
\else
\fi

\usepackage{amsmath}
\usepackage{amssymb}
\usepackage{url}
\usepackage{amsmath,amsfonts,bm}

\def\eqref#1{equation~\ref{#1}}
\def\floor#1{\lfloor #1 \rfloor}
\def\1{\bm{1}}

\def\vx{{\bm{x}}}
\def\vy{{\bm{y}}}

\def\mI{{\bm{I}}}

\DeclareMathAlphabet{\mathsfit}{\encodingdefault}{\sfdefault}{m}{sl}
\SetMathAlphabet{\mathsfit}{bold}{\encodingdefault}{\sfdefault}{bx}{n}

\newcommand{\norm}[1]{\left\lVert#1\right\rVert}
\newcommand{\E}{\mathbb{E}}
\newcommand{\Ls}{\mathcal{L}}

\usepackage[breaklinks=true,bookmarks=false]{hyperref}

\begin{document}

\title{Fast Inference in Denoising Diffusion Models via MMD Finetuning}

\author{Emanuele~Aiello, Diego~Valsesia, and~Enrico~Magli% <-this % stops a space
\thanks{The authors are with Politecnico di Torino -- Department of Electronics and Telecommunications, Italy. email: \{name.surname\}@polito.it.}% <-this % stops a space
}

\maketitle

\begin{abstract}
Denoising Diffusion Models (DDMs) have become a popular tool for generating high-quality samples from complex data distributions. These models are able to capture sophisticated patterns and structures in the data, and can generate samples that are highly diverse and representative of the underlying distribution. However, one of the main limitations of diffusion models is the complexity of sample generation, since a large number of inference timesteps is required to faithfully capture the data distribution.
In this paper, we present MMD-DDM, a novel method for fast sampling of diffusion models. Our approach is based on the idea of using the Maximum Mean Discrepancy (MMD) to finetune the learned distribution with a given budget of timesteps. This allows the finetuned model to significantly improve the speed-quality trade-off, by substantially increasing fidelity in inference regimes with few steps or, equivalently, by reducing the required number of steps to reach a target fidelity, thus paving the way for a more practical adoption of diffusion models in a wide range of applications.
We evaluate our approach on unconditional image generation with extensive experiments across the CIFAR-10, CelebA, ImageNet and LSUN-Church datasets. Our findings show that the proposed method is able to produce high-quality samples in a fraction of the time required by widely-used diffusion models, and outperforms state-of-the-art techniques for accelerated sampling. Code is available at: \url{https://github.com/diegovalsesia/MMD-DDM}.
\end{abstract}

\begin{figure*}[t!]
\vspace{-0.4cm}
\begin{center}
{
\begin{tabular}{@{}c@{\hspace{.1cm}}c@{\hspace{.1cm}}c@{\hspace{.1cm}}c@{\hspace{.35cm}}c@{\hspace{.1cm}}c@{}}
      & DDIM & MMD-DDM (Inception-V3) & MMD-DDM (CLIP) & Reference &   \\
     
     \raisebox{.2cm}{\rotatebox{90}{CelebA}} & 
     \makebox{\includegraphics[width=.225\linewidth]{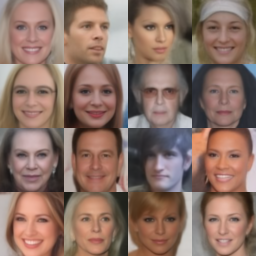}} & 
     \makebox{\includegraphics[width=.225\linewidth]{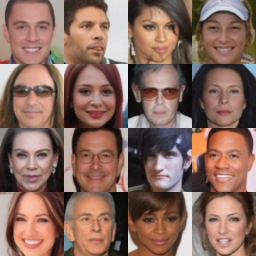}} & 
     \makebox{\includegraphics[width=.225\linewidth]{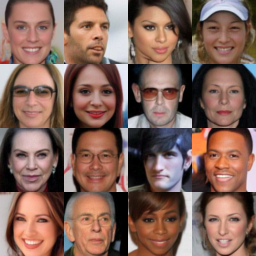}} &
     \makebox{\includegraphics[width=.225\linewidth]{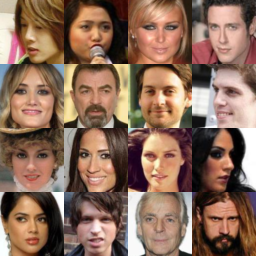}} \\     
     \raisebox{.6cm}{\rotatebox{90}{\textit{CIFAR-10}}} & 
     \makebox{\includegraphics[width=.225\linewidth]{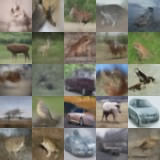}} & 
     \makebox{\includegraphics[width=.225\linewidth]{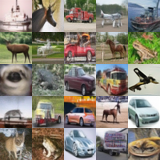}} & 
    \makebox{\includegraphics[width=.225\linewidth]{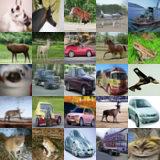}} &
    \makebox{\includegraphics[width=.225\linewidth]{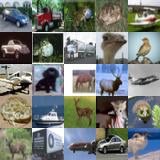}} \\
\end{tabular}
}
\end{center}

\caption{Generated samples for CelebA (top) and CIFAR-10 (bottom). The samples are obtained using 5 timesteps with the DDIM sampling procedure. Results from standard DDIM (left), the same model finetuned using MMD with Inception-V3 features (center-left) and CLIP features (center-right), reference images from the dataset (right). Samples are not cherry-picked. Finetuning improves details clarity and sharpness, occasionally introducing semantic changes.
\label{fig:samples}}
\vspace{-0.1cm}
\end{figure*}

\section{Introduction}

Denoising Diffusion Models (DDMs) \cite{sohl2015deep}\cite{ho2020denoising}\cite{song2019generative} have emerged as a powerful class of generative models. DDMs learn to reverse a gradual multi-step noising process to match a data distribution. Samples are then produced by a Markov Chain that starts from white noise and progressively denoises it into an image. This class of models has shown excellent capabilities in synthesising high-quality images \cite{nichol2021improved}\cite{rombach2022high}, audio \cite{chen2020wavegrad}, and 3D shapes \cite{zhou20213d} \cite{zenglion}, recently outperforming Generative Adversarial Networks (GANs) \cite{goodfellow2014generative} on image synthesis.

However, GANs require a single forward pass to generate samples, while the iterative DDM design requires hundreds or thousands of inference timesteps and, consequently, forward passes through a denoising neural network.
The slow sampling process thus represents one of the most significant limitations of DDMs. It is well-known that there is a trade-off between sample quality and speed, measured in the number of timesteps \cite{song2020denoising}\cite{nichol2021improved}. However, it is currently unclear how low the number of timesteps can be pushed while retaining high quality for a given data distribution \cite{chen2020wavegrad}.
This issue is the focus of a lot of current research in the field, with recent works proposing acceleration solutions which can be divided into two categories: learning-free sampling and learning-based sampling.
The learning-free approach focuses on modifying the sampling process without the need for training \cite{song2020denoising}\cite{song2020score}\cite{jolicoeur2021gotta}\cite{karras2022elucidating}, while the learning-based approach uses techniques such as truncation \cite{lyu2022accelerating}\cite{zheng2022truncated}, knowledge distillation \cite{salimans2021progressive}\cite{luhman2021knowledge}, dynamic programming \cite{watson2021learning} and differentiable sampler search \cite{watson2022learning} to improve sampling speed. 

In this paper, we propose MMD-DDM, a technique to finetune a pretrained DDM with a large number of timesteps in order to optimize the features of the generated data under the constraint of a reduced number of timesteps. This is done by directly optimizing the weights of the denoising neural network via backpropagation through the sampling chain. The minimization objective is the Maximum Mean Discrepancy (MMD) \cite{gretton2006kernel} between real and generated samples in a perceptually-relevant feature space. This allows to specialize the model for a fixed and reduced computational budget with respect to the original training; the use of MMD represents a different and, possibly complementary, objective to the original denoising loss. Our proposed approach is extremely fast, requiring only a small number of finetuning iterations. Indeed, the finetuning procedure can be performed in minutes, or at most few hours for more complex datasets, on standard hardware. Moreover, it is agnostic to the sampling procedure, making it appealing even for future models employing new and improved procedures. 

Extensive experimental evaluation suggests that the proposed solution significantly outperforms state-of-the-art approaches for fast DDM inference. MMD-DDM is able to substantially reduce the number of timesteps required to reach a target fidelity. We also need to remark that the choice of feature space for the MMD objective may artificially skew results, if the evaluation metric is based on the same feature space being optimized, such as Inception features and the FID score. We discuss the importance of this point for fair evaluation and present results on different metrics and features spaces, such as CLIP features \cite{radford2021learning}, in order to present a fair assessment of the method.

%------------------------------------------------------------------------------------------
\section{Background and Related Work}
\subsection{Denoising Diffusion Models}
DDMs (\cite{ho2020denoising}\cite{sohl2015deep}) leverage the diffusion process to model a specific distribution starting from random noise. They are based on a predefined Markovian forward process, by which data are progressively noised in $T$ steps. $T$ is set to be sufficiently large such that $\vx_T$ is close to white Gaussian noise (in practice, $T \geq 1000$  is often used). The forward process can be written as:
\begin{eqnarray}
\label{eqn:fwd}
    q(\vx_0,...,\vx_T) &=& q(\vx_0)\prod_{t=1}^T q(\vx_t|\vx_{t-1}) \\
    q(\vx_t|\vx_{t-1}) &=&  \mathcal{N}(\vx_t|\sqrt{1-\beta_t}\vx_s, \beta_t \mI)    
\end{eqnarray}
where $q(\vx_0)$ denotes the real data distribution and $\beta_t$ the variance of the Gaussian noise at timestep $t$.
The reverse process traverses the Markov Chain backwards and can be written as:
\begin{eqnarray}
    p_\theta(\vx_{0:T}) &=& p(\vx_T) \prod\nolimits_{t=1}^T p_\theta(\vx_{t-1} \mid \vx_t) \\
    p_\theta(\vx_{t-1} \mid \vx_{t}) &=& \mathcal{N}(\vx_{t-1} \mid \mu_{\theta}({\vx}_{t}, t), \sigma_t^2\bm{I})~. \label{eq:reverse_process}
\end{eqnarray}
The parameters of the learned reverse process $p_\theta$ can be optimized by maximizing an evidence lower bound (ELBO) on the training set.
Under a specific parametrization choice \cite{ho2020denoising}, the training objective can be simplified to that of a noise conditional score network \cite{vincent2011connection} \cite{song2019generative}: 
\begin{align}
\min_\theta \Ls(\theta) = \E_{x_0, \epsilon, t} || \epsilon - \epsilon_{\theta}(\vx_t, t)||^2_2
\end{align}
where $\vx_0 \sim q_\text{data}$, $\epsilon \sim \mathcal{N}(0, \bm{I})$ and $t$ is uniformly sampled from \{$1,...,T$\}.

\subsection{Accelerated Sampling for DDMs}\label{sec:related}

Accelerated DDM sampling is currently a hot research topic. At a high level, the different approaches can be divided into two categories: learning-free sampling and learning-based sampling \cite{yang2022diffusion}. The learning-free approaches do not require training and instead focus on modifying the sampling process to make it more efficient. One example is the work of Song et al. \cite{song2020denoising} (DDIM), in which they define a new family of non-Markovian diffusion processes that maintains the same training objectives as a traditional DDPM. They demonstrate that alternative ELBOs may be built using only a sub-sequence of the original timesteps $\tau \in$ \{$1,...,T$\}, obtaining faster samplers compatible with a pre-trained DDPM.
Other works focus on using the Score SDE formulation \cite{song2020score} of continuous-time DDMs to develop faster sampling methods. For example, Song et al. \cite{song2020score} propose the use of higher-order solvers such as Runge-Kutta methods, while Jolicoeur et al. \cite{jolicoeur2021gotta} propose the use of SDE solvers with adaptive timestep sizes. Another approach is to solve the probability flow ODE, which has been shown by Karras et al. \cite{karras2022elucidating} to provide a good balance between sample quality and sampling speed when using Heun's second-order method. Additionally, customized ODE solvers such as the DPM-solver \cite{lu2022dpm} and the Diffusion Exponential Integrator sampler \cite{zhang2022fast} have been developed specifically for DDMs and have been shown to be more efficient than general solvers. These methods provide efficient and effective ways to speed up the sampling process in continuous-time DDMs. 

The other main line of approaches for efficient sampling is the learning-based one. Some of these approaches \cite{lyu2022accelerating}\cite{zheng2022truncated} involve truncating the forward and reverse diffusion processes to improve sampling speed, while others \cite{salimans2021progressive} \cite{luhman2021knowledge} use knowledge distillation to create a faster model that requires fewer steps. Another approach (GENIE) \cite{dockhorngenie}, based on truncated Taylor methods, trains an additional model on top of a first-order score network to create a second-order solver that produces better samples with fewer steps. Dynamic programming techniques \cite{watson2021learning} have also been used to find the optimal discretization scheme for DDMs by selecting the best time steps to maximize the training objective, although the variational lower bound does not correlate well with sample quality, limiting the performance of the method. In a successive work \cite{watson2022learning}, the sampling procedure was directly optimized using a common perceptual evaluation metric (KID) \cite{binkowski2018demystifying}, but this required a long training time (30k training iterations).
In this work. the authors backpropagate through the sampling chain using reparametrization and gradient rematerialization in order to make the optimization feasible. Our work is closely related to \cite{watson2022learning}, since we similarly backpropagate through the sampling chain. However, we use the MMD \cite{gretton2006kernel}\cite{gretton2012kernel} to finetune the weights of a pretrained DDM without optimizing the sampling strategy. In essence, the proposed method is complementary to \cite{watson2022learning}: instead of optimizing the sampling procedure, keeping the model fixed, we directly optimize the model leaving the sampling procedure unchanged. This leads to better results with as few as 500 finetuning iterations. We also remark that our approach is decoupled from the sampling strategy and can be used in conjunction with other training-free acceleration methods such as DDIM.

\subsection{MMD in Generative Models}
The MMD \cite{gretton2006kernel}\cite{gretton2012kernel} is a distance on the space of probability measures. It is a non-parametric approach that does not make any assumptions about the underlying distributions, and can be used to compare a wide range of distributions. Generative models trained by minimizing the MMD were first considered in \cite{pmlr-v37-li15}\cite{dziugaite2015training}. These works optimized a generator to minimize the MMD with a fixed kernel, but struggled with the complex distribution of natural images where pixel distances are of little value. Successive works \cite{li2017mmd} \cite{binkowski2018demystifying} addressed this problem by adversarially learning the kernel for the MMD loss, reaching results comparable to GANs trained with a Wasserstein critic. In this work we apply the MMD in the context of diffusion models, demonstrating its effectiveness in finetuning a pretrained DDM under a more restrictive timesteps constraint.

%-----------------------------------------------------------------------------------------
\section{Method}

\subsection{Overview}
We propose MMD-DDM, a technique to accelerate inference in DDMs while maintaining high sample quality, based on finetuning a pretrained diffusion model. The finetuning process minimizes an unbiased estimator of the MMD between real and generated samples, evaluated over a perceptually-relevant feature space. We backpropagate through the sampling process with the aid of the reparametrization trick and gradient checkpointing. This is done only for a small subset of the original timesteps and can be combined with existing techniques for timestep selection or acceleration of the sampling process.
The reduction in timesteps with respect to the original model degrades the distribution of the generated data. However, the main idea behind the proposed approach is that it is possible to recover part of this degradation by analyzing the generated data in a perceptual feature space and imposing that the reduced DDM produces perceptual features similar to those of real data via MMD minimization. 
By utilizing this approach, we are thus able to maximize the model performance under a fixed computational budget. It is interesting to notice that older approaches that utilized MMD as sole objective for image generation failed to capture their complex data distribution. On the other hand, our approach avoids that as it leverages the strong baseline provided by the pretrained DDM, albeit degraded by the timesteps constraint.

\subsection{Finetuning with MMD}
We are interested in learning a model distribution $p_{\theta}(\vx_0)$ that approximates the real data distribution $q(\vx_0)$. Starting from a pretrained diffusion model, we know from previous work (DDIM \cite{song2020denoising}) that it is possible to sample from $p_{\theta}^{(\mathcal{T})}(\vx_0)$, i.e., the learned distribution using a subset of the original timesteps $\mathcal{T} \subset $ \{$1,...,T$\}, accepting a complexity-quality tradeoff.
The MMD \cite{gretton2006kernel} is an integral probability metric that we use to measure the discrepancy between the real data distribution $q(\vx_0)$ and the generated data distribution with the given budget of timesteps $p_{\theta}^{(\mathcal{T})}(\vx_0)$. Mathematically, it is defined as: 
\begin{align}
\label{eq:mmdprimal}
\centering
\mathrm{MMD}(p_\theta^{(\mathcal{T})},q) = \Vert \E_{\vx \sim p_\theta^{(\mathcal{T})}} \varphi(\vx) - \E_{\vy \sim q}\varphi(\vy) \Vert
%    &\mathcal{L}_{\mathrm{MMD}^2} = \left \|\frac{1}{N} \sum_{i=1}^N \phi(\vx_i) -
%    \frac{1}{M} \sum_{j=1}^M \phi(\vy_j)\right \|^2  
\end{align}
where $\varphi$ represents a function mapping raw images to a perceptually-meaningful feature space. This is needed as MMD would not perform well on the pixel space, since it is well known that images live on a low-dimensional manifold within the high-dimensional pixel space. However, once the images are mapped into an appropriate feature space, MMD is proven to have strong discriminative performances, as proved by the success of the KID \cite{binkowski2018demystifying} as evaluation metric for perceptual quality. The choice of feature space is critical for the performance of the proposed method and for the fair assessment of methods optimizing quality metrics, which will be presented in Secs. \ref{sec:feature_spaces} and \ref{sec:feature}.

In order to use the MMD as our loss function, given a batch of generated samples $\{{\vx_i}\}_{i=1}^N \sim p_{\theta}^{(\mathcal{T})}(\vx_0)$ and a batch of real samples $\{{\vy_i}\}_{i=1}^N \sim q(\vx_0)$, we use the unbiased estimator proposed by Gretton et al. \cite{gretton2012kernel}: 
\begin{align}
\centering
    \mathcal{L}_{\mathrm{MMD}^2}^\text{unbiased} =  &\frac{1}{N(N-1)}\sum_{i\neq j}^n k(\phi(\vx_i),\phi(\vx_j)) \nonumber  \\ 
    & -\frac{2}{N^2}\sum_{i=1}^N\sum_{j=1}^N k(\phi(\vx_i),\phi(\vy_j)) + c.
\end{align}
where $N$ is the batch size, $c$ is a constant, and $k$ is a generic positive definite kernel (in our experiments we consider linear, cubic and Gaussian kernels, see Sec. \ref{sec:ablations}). The loss function is minimized in order to finetune the values of the parameters $\theta$ of a pretrained denoising neural network composing the diffusion model. Next, we are going to discuss the choice of the feature extraction function $\phi$. 

\begin{table*}[t]
    \caption{Unconditional CIFAR-10 generative performance (Inception FID).}
    %\vspace{10pt}
    \label{cifar-res}
    \begin{center}
        \begin{tabular}{l c c c c}
        \hline
        Method &  $\vert \mathcal{T} \vert=5$ & $\vert \mathcal{T} \vert=10$ & $\vert \mathcal{T} \vert=15$ & $\vert \mathcal{T} \vert=20$  \\%& NFEs=25 \\ %
        \hline\hline
        DDPM \cite{ho2020denoising}& 76.3 & 42.1 & 31.4 & 25.9 \\
        DDIM~\cite{song2020denoising} & 32.7 & 13.6 & 9.31 & 7.50 \\
        \textbf{DDIM + MMD-DDM} (Inception-V3) &  \textbf{5.48}  &  \textbf{3.80}  &  \textbf{4.11}  &    \textbf{3.55}  \\
        \textbf{DDIM + MMD-DDM} (CLIP) & 6.79  & 4.87 & 4.79 & 4.52 \\
        \hline
        GENIE \cite{dockhorngenie} & 13.9 & 5.97 & 4.49 & 3.94\\% & 3.67 \\ %
        PNDM~\cite{liu2022pseudo} & 35.9 & 10.3 & 6.61 & 5.20 \\%& 4.51 \\%& 3.30 \\
        FastDPM~\cite{kong2021fast}  & - & 9.90 & - & 5.05\\% & - \\
        Learned Sampler~\cite{watson2022learning}  & 13.8 & 8.22 & 6.12 & 4.72 \\%& 4.25 
        Analytic DDIM~\cite{bao2022analyticdpm} & - & 14.7 & 9.16 & 7.20\\% & 5.71 &\\% 4.04 \\
        DPM-Solver(Type-1) \cite{lu2022dpm} & - & 6.37 & 3.78 & 4.28 \\
        DPM-Solver(Type-2) \cite{lu2022dpm} & - & 10.2 & 4.17 & 3.72 \\
        \hline
    \end{tabular}
    \end{center}
\end{table*}

\begin{table*}[t]
    \caption{Unconditional CelebA generative performance (Inception FID).}
    \label{celeb-res}
    \begin{center}
        \begin{tabular}{l c c c c}
        \hline
        Method &  $\vert \mathcal{T} \vert=5$ & $\vert \mathcal{T} \vert=10$ & $\vert \mathcal{T} \vert=15$ & $\vert \mathcal{T} \vert=20$  \\%& NFEs=25 \\ %
        \hline\hline
        DDIM~\cite{song2020denoising} & 22.4 & 17.3 & 16.0 & 13.7  \\
        \textbf{DDIM + MMD-DDM} (Inception-V3) & \textbf{3.04}   & \textbf{2.58}   & \textbf{2.13}   &  \textbf{2.24}  \\
        \textbf{DDIM + MMD-DDM} (CLIP) & 4.65 & 3.90 & 3.17 & 3.27 \\
        \hline
        ES+StyleGAN2+DDIM \cite{lyu2022accelerating} & 9.15 & 6.44 & - & 4.90 \\
        PNDM~\cite{liu2022pseudo} & 11.3 & 7.71 & - & 5.51 \\%& 4.51 \\%& 3.30 \\
        FastDPM~\cite{kong2021fast}  & - & 15.3 & - & 10.7\\% & - \\
        Diffusion Autoencoder \cite{preechakul2022diffusion} & - & 12.9 & - & 10.2 \\
        Analytic DDPM \cite{bao2022analyticdpm} & - & 29.0 & 21.8 & 18.1 \\
        Analytic DDIM \cite{bao2022analyticdpm} & - & 15.6 & 12.3 & 10.45 \\
        DPM-Solver(Type-1) \cite{lu2022dpm} & - & 6.92 & 3.05 & 2.82 \\
        DPM-Solver(Type-2) \cite{lu2022dpm} & - & 5.83 & 3.11 & 3.13\\
        \hline
    \end{tabular}
    
    \end{center}
\end{table*}

\begin{table}[t]
    \caption{Unconditional ImageNet generative performance (Inception FID).}
    %\vspace{10pt}
    \label{imagenet-res}
    \setlength\tabcolsep{1pt} 
    \begin{center}
        \begin{tabular}{l c c c }
        \hline
        Method &  $\vert \mathcal{T} \vert=5$ & $\vert \mathcal{T} \vert=10$ & $\vert \mathcal{T} \vert=20$  \\
        \hline\hline
        DDIM~\cite{song2020denoising} & 131.5 & 35.2 & 20.7  \\
        \textbf{DDIM + MMD-DDM} (Inc-V3) &  33.1  & 21.1   &   \textbf{12.4}   \\
        \textbf{DDIM + MMD-DDM} (CLIP) & \textbf{27.5} & \textbf{16.4} & 14.5  \\
        \hline
       Learned Sampler \cite{watson2022learning} & 55.1 & 37.2 & 24.6\\
       Analytic-DDIM \cite{bao2022analyticdpm} & - & 70.6 & 30.9 \\
       Analytic-DDPM \cite{bao2022analyticdpm} & - & 60.6 & 37.7 \\
       DPM-Solver(T2) \cite{lu2022dpm} & - & 24.4 & 18.53\\
        \hline
    \end{tabular}
        
    \end{center}
\end{table}

\begin{table}[t]
    \caption{Unconditional LSUN-Church Outdoor generative performance (Inception FID).}
    %\vspace{10pt}
    \label{lsun-res}
    \setlength\tabcolsep{1pt} 
    \begin{center}
        \begin{tabular}{l c c c c}
        \hline
        Method &  $\vert \mathcal{T} \vert=5$ & $\vert \mathcal{T} \vert=10$ & $\vert \mathcal{T} \vert=20$  \\
        \hline\hline
        DDIM~\cite{song2020denoising} & 49.6 & 19.4 & 12.5   \\
        \textbf{DDIM + MMD-DDM} (Inc-V3) & \textbf{4.75} &  \textbf{7.55}  &  \textbf{6.21}  \\  
        \textbf{DDIM + MMD-DDM} (CLIP) & 14.2  & 10.7 &  8.82  \\
        \hline
        S-PNDM \cite{liu2022pseudo} & 20.5 & 11.8 & 9.20 \\
        F-PNDM \cite{liu2022pseudo} & 14.8 & 8.69 & 9.13 \\

        \hline
    \end{tabular}
    
    \end{center}
\end{table}

\subsection{Perceptually-Relevant Feature Spaces}
\label{sec:feature_spaces}

As we previously mentioned, it is necessary to embed real and generated images in some perceptually-relevant feature space, so that the MMD objective could be effective. The feature mapping network $\phi$ plays a crucial role in the performance of the method. However, this is not a trivial choice. The most popular choice could be to use the feature space of the penultimate layer of an ImageNet-pretrained Inception-V3 classifier \cite{szegedy2016rethinking}. This choice is widely used to evaluate performance of generative models, with Inception Score (IS) \cite{salimans2016improved}, FID \cite{heusel2017gans} and KID \cite{binkowski2018demystifying} all using it. 

However, a recent study \cite{kynkaanniemi2022role} has examined the effectiveness of using ImageNet-pretrained representations to evaluate generative models, and found that the presence of ImageNet classes has a significant impact on the evaluation. The study highlights some potential pitfalls in using these metrics, and how they can be manipulated by the use of ImageNet pretraining. This suggests that care should be taken when using ImageNet features to optimize generative models as this can potentially distort the FID quality metric and make it unreliable. Indeed, for any image generation method, part of the improvement might lie in the \textit{perceptual null space} \cite{kynkaanniemi2022role} of FID, which encompasses all the operations that change the FID without affecting the generated images in a perceptible way. For our finetuning procedure, we have experimentally observed a better overall visual quality of generated images and a consistent gain in FID, when optimizing MMD with Inception features. However, it is hard to quantitatively assess how much of this improvement is due to actual perceptual improvements versus optimizations in the perceptual null space. These considerations apply also to the work of Watson et al. \cite{watson2022learning}. 

One solution to this problem is to use a different feature space for the feature mapping network, such as one that has not been pretrained on ImageNet. Thus, we propose to optimize the MMD using the feature space of the CLIP image encoder \cite{radford2021learning}, which has been trained in a self-supervised way and is supposed to have richer representations without exposure to ImageNet classes. Moreover, we also consider the case in which we optimize MMD with Inception features and measure performance with a variant of FID using CLIP features. 
More comments, details, and a discussion of the various results can be found in Sec. \ref{sec:feature}.

\begin{figure*}[t!]
\label{figure:res2}
\vspace{-0.4cm}
%\small
\begin{center}
{%\small
\begin{tabular}{@{}c@{\hspace{.1cm}}c@{\hspace{.1cm}}c@{\hspace{.1cm}}c@{\hspace{.35cm}}c@{\hspace{.1cm}}c@{}}
      & DDIM & MMD-DDM (Inception-V3) & MMD-DDM (CLIP) & Reference &   \\
     
     \raisebox{.2cm}{\rotatebox{90}{\textit{LSUN-Church}}} & 
     \makebox{\includegraphics[width=.225\linewidth]{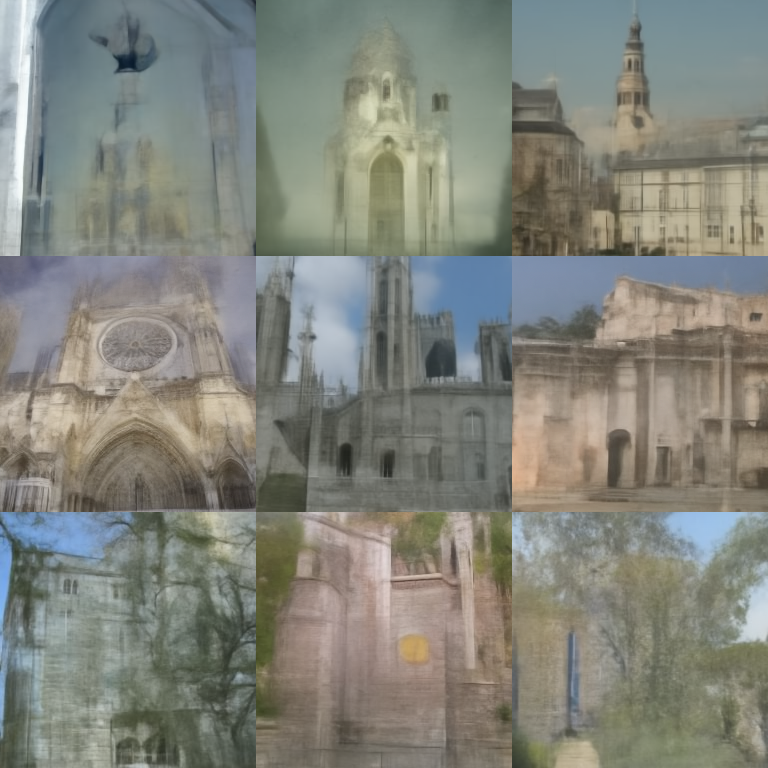}} & 
     \makebox{\includegraphics[width=.225\linewidth]{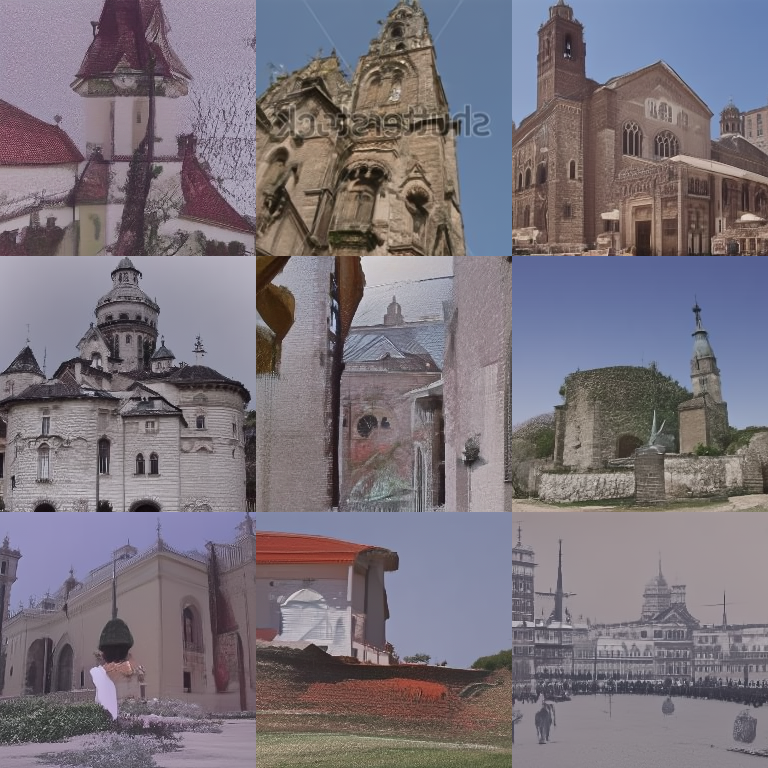}} & 
     \makebox{\includegraphics[width=.225\linewidth]{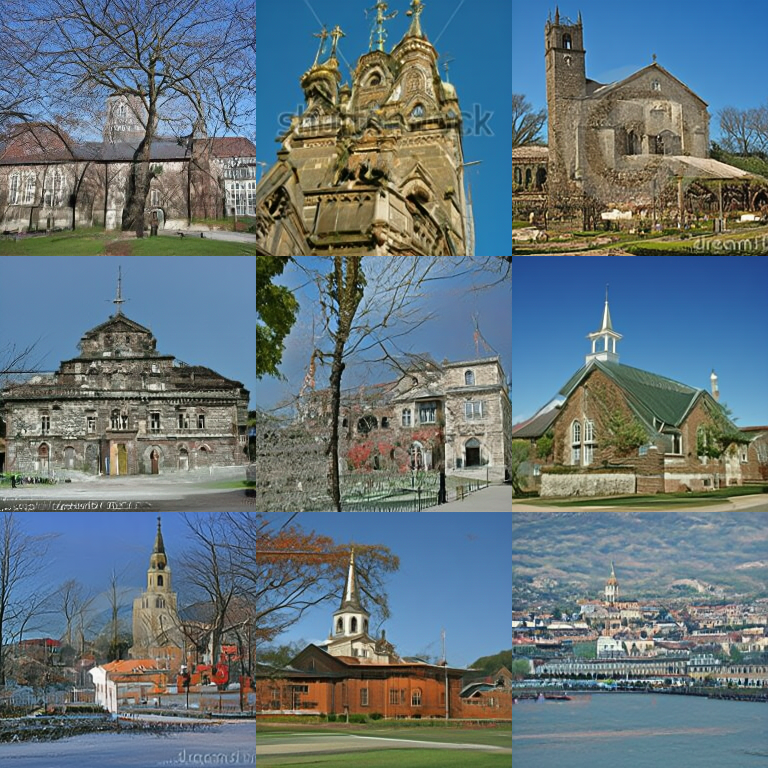}} &
     \makebox{\includegraphics[width=.225\linewidth]{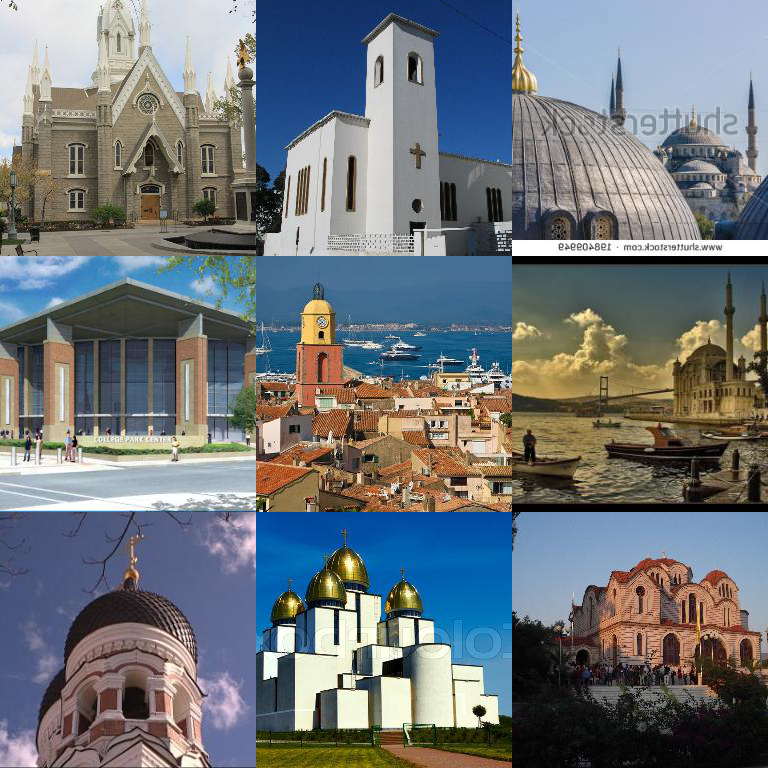}} \\     
     \raisebox{.6cm}{\rotatebox{90}{\textit{ImageNet}}} & 
     \makebox{\includegraphics[width=.225\linewidth]{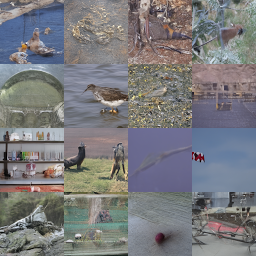}} & 
     \makebox{\includegraphics[width=.225\linewidth]{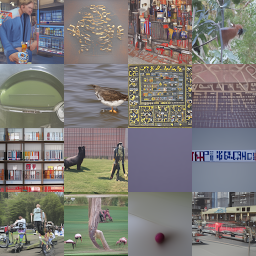}} & 
    \makebox{\includegraphics[width=.225\linewidth]{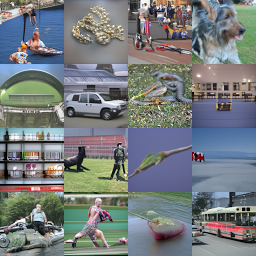}} &
    \makebox{\includegraphics[width=.225\linewidth]{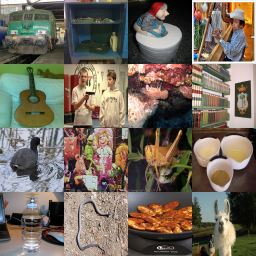}} \\
\end{tabular}
}
\end{center}

\caption{Generated samples for LSUN-Church (top) and ImageNet (bottom). The samples are obtained using 5 timesteps for LSUN-Church and 10 timesteps for ImageNet, with the DDIM sampling procedure. Results from Standard DDIM (left), the same model finetuned using Inception-V3 features (center-left) and CLIP features (center-right), reference images from the dataset (right). Samples are not cherry-picked.
\label{fig:samples2}}
\vspace{-0.1cm}
\end{figure*}

%--------------------------------------------------------------------------------------
\section{Experiments}

\subsection{Setting}
\paragraph{Datasets} In order to demonstrate the effectiveness of the proposed solution, we validate it on several datasets with different resolutions. We use CIFAR-10 \cite{krizhevsky2009learning} at resolution $32\times32$, CelebA \cite{liu2015faceattributes} at resolution $64\times64$, Image-Net \cite{IMAGENET} at resolution $64\times64$, and LSUN-Church \cite{yu2015lsun}  at resolution $256\times256$.

\paragraph{Models and Sampling} We use the models pretrained by Ho et al. \cite{ho2020denoising} for the CIFAR-10 and LSUN experiments, the model pretrained by Song et al. \cite{song2020denoising} for CelebA, and the  model pretrained by Nichol and Dhariwal \cite{nichol2021improved} with the $L_\text{hybrid}$ objective for ImageNet. All the architectures are based on the modified UNet \cite{ronneberger2015u} that incorporates self-attention layers \cite{vaswani2017attention}. We perform our experiments using the efficient sampling strategy of DDIM \cite{song2020denoising}, as it already has good performance in few-timesteps regime. We fix the timestep schedule in the main experiments to be linear. The MMD kernel is polynomial cubic in all experiments, except the kernel ablation one.
We also test the proposed solution with the DDPM \cite{ho2020denoising} sampling strategy in Sec. \ref{sec:ablations}.

\paragraph{Evaluation} We use the FID \cite{heusel2017gans} to evaluate sample quality. All the values are evaluated by comparing 50k real and generated samples as this is the literature's standard. We also use FID$_\text{CLIP}$ \cite{kynkaanniemi2022role} in some experiments to remove the effect of Image-Net classes in the evaluation. Additional evaluation metrics such as Inception Score \cite{salimans2016improved}, Spatial FID \cite{nash2021generating}, and Precision and Recall \cite{sajjadi2018assessing} can be found in the Supplementary Material.

\paragraph{Implementation Details} For all the experiments we set the batch size equal to 128. We use Adam as optimizer \cite{kingma2015adam} with $\beta_1 = 0.9$, $\beta_2 = 0.999$, $\epsilon = 1 \times 10^{-8}$ and learning rate equal to $5\times 10^{-6}$. When DDIM is used, we set $\sigma_t=0$. As feature extractors, we use the standard Inception-V3\footnote{http://download.tensorflow.org/models/image/imagenet/inception-2015-12-05.tgz} pretrained on Image-Net, and the ViT-B/32\footnote{https://github.com/openai/CLIP} model from CLIP \cite{radford2021learning}. We use \textit{torch-fidelity} \cite{obukhov2020torchfidelity} for the FID evaluation. We train all the models for about 500 iterations. Finetuning with a budget of 5 timesteps required about 10 minutes for CIFAR-10, about 45 minutes for CelebA, and about one hour for ImageNet on a single Nvidia RTX A6000. For LSUN-Church and for the other timesteps budgets, finetuning has been performed on four Nvidia RTX A6000. Finetuning for 5 timesteps of LSUN-Church required about two hours on the mentioned hardware.

\begin{table*}[t!]
    \caption{Comparison of relative improvements evaluating FID in Inception-V3 feature space versus CLIP feature space.}
    \vspace{10pt}
    \label{fid_clip}
    \begin{center}
       \begin{tabular}{l c c c c}
    \hline
     & \multicolumn{2}{c}{$\vert \mathcal{T} \vert=5$} & \multicolumn{2}{c}{$\vert \mathcal{T} \vert=10$} \\\hline \hline
    \multicolumn{5}{c}{\textit{CIFAR-10}} \\ \hline
      & FID & FID$_\text{CLIP}$ & FID & FID$_\text{CLIP}$ \\
    \hline     
     DDIM & 32.7 & 13.7 & 13.6 & 6.87 \\
     \textbf{DDIM + MMD-DDM} (Inception-V3) & 5.48 & 2.11 & 3.80 & 2.01 \\
     Improvement & -83.2\%  & -84.4\% & -72.0\% & -70.7\% \\     
    \hline \hline
     \multicolumn{5}{c}{\textit{CelebA}} \\\hline
     & FID & FID$_\text{CLIP}$ & FID & FID$_\text{CLIP}$ \\\hline
     DDIM & 22.4 & 12.2 & 17.3 & 9.48 \\
     \textbf{DDIM + MMD-DDM} (Inception-V3) & 3.04 & 4.94 & 2.58 & 4.26  \\
     Improvement & -86.4\% & -59.3\%  & -85.0\% & -55.0\%  \\ 
     \hline \hline
    \multicolumn{5}{c}{\textit{ImageNet}} \\\hline
     & FID & FID$_\text{CLIP}$ & FID & FID$_\text{CLIP}$ \\\hline
     DDIM & 131.5 & 29.5 & 35.2 & 11.9  \\
     \textbf{DDIM + MMD-DDM} (Inception-V3) & 33.1 & 15.2 &  21.1 & 9.21   \\
     Improvement & -74.8\% & -56.8\% & -40.0\% &  -22.6\% \\ 
     \hline
        \end{tabular}
    \end{center}
\end{table*}
\subsection{Image Generation Results}

We evaluate MMD-DDM using the following timesteps budgets: $\vert \mathcal{T} \vert \in \{5, 10, 15, 20\}$. We report the values of FID on unconditional generation experiments for CIFAR-10 in Table \ref{cifar-res}, for CelebA in Table \ref{celeb-res}, for ImageNet in Table \ref{imagenet-res} and for LSUN-Church in Table \ref{lsun-res}.

We compare against several state-of-the-art methods for accelerating DDMs. The tables report the results for MMD-DDM trained with Inception features and we also report results taken from literature for other methods. 
For all datasets and timesteps budgets, MMD-DDM provides superior or, occasionally, comparable quality to state-of-the art approaches.
For the ImageNet experiment, we remark that we report the result of the Learned Sampler approach \cite{watson2022learning}, which uses an improved version of the model from \cite{nichol2021improved} trained for 3M iterations, instead of the 1.5M iterations used by our checkpoint, thus making the comparison slightly unfavourable for our method.
We do not compare with the progressive distillation method \cite{salimans2021progressive}, as it cannot be considered a post-training acceleration technique but rather a very computationally-demanding modification of the DDM training procedure.

Qualitative comparisons for CIFAR-10 and CelebA are shown in Fig. \ref{fig:samples} and for LSUN-Church and ImageNet in Fig. \ref{fig:samples2}. More generated samples, for different numbers of timesteps, can be found in the Supplementary Material. It can be noticed that MMD-DDM provides substantial improvements in visual quality when the number of timesteps is highly constrained. As the available timesteps budget is relaxed to 20 or more, the improvement provided MMD-DDM diminishes, although all approaches start providing high quality samples.

\subsection{Feature Space Discussion}\label{sec:feature}

Results in the previous section were presented with the commonly-used FID metric exploiting Inception features. However, as detailed in Sec. \ref{sec:feature}, our optimization of Inception features via the MMD loss could raise concerns about the reliability of the FID metric. In this section, we present results using the CLIP feature space in either the MMD loss or the FID metric.

Figs. \ref{fig:samples} and \ref{fig:samples2} already show a visual comparison between using the MMD with Inception features and CLIP features and more results are present in the Supplementary Material. It can be noticed that optimizing over CLIP features leads to higher visual quality, including sharper details and clarity, confirming that the CLIP space is a superior embedding of perceptually-relevant features. As a reference, we also report the FID scores obtained by MMD-DDM with CLIP features in Tables \ref{cifar-res},\ref{celeb-res},\ref{imagenet-res},\ref{lsun-res}. Notice that lower values are observed, possibly due to the reliance of the FID on flawed Inception features and the metric not accurately tracking a genuine improvement in visual quality. 

We further expand the set of results in Table \ref{fid_clip}, in which we optimize Inception features with the MMD but then measure quality using the FID computed on CLIP features (FID$_\text{CLIP}$), as proposed in \cite{kynkaanniemi2022role}. Since the feature space used for evaluation is different from the one used in optimization, the observed gains in FID$_\text{CLIP}$ suggest us that the quality improvement in quantitatively meaningful and not just an artifact of the metric. Percentage improvements in FID$_\text{CLIP}$ mostly track those of regular FID, albeit being lower in some cases, suggesting that some overfitting of the perceptual null space does indeed happen when Inception features are used for both MMD and FID.

\subsection{Analysis of Overfitting}

\begin{figure}
\centering
\includegraphics[width=0.95\linewidth]{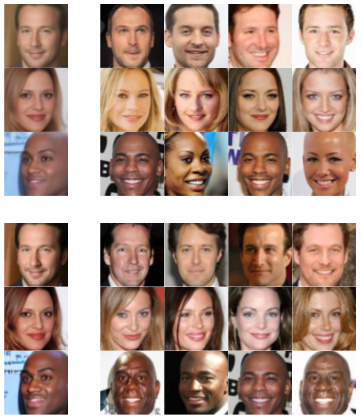}\\
\caption{Generated samples by the DDIM model (top) and the finetuned model (bottom) for CelebA. For each generated samples we visualize the top-4 nearest neighbours.}
\label{figure:NN}
\end{figure}

One might wonder whether finetuning via the MMD loss leads to images overfitting the features of the training set. This section presents a experiment to dispel this concern.
To do so, we looked at the top-$K$ nearest neighbors of generated samples when the CLIP feature space is used for both the optimization with the MMD and the space for nearest neighbor search (Euclidean distance betweeen CLIP features). 
Fig. \ref{figure:NN} provides the results of this experiment for samples generated with both the pretrained model and the finetuned model. 
We can see that the nearest neighbors of the samples generated after finetuning are not more significantly similar to the generated image than those for the pretrained model. More samples can be found in the Supplementary material.

\subsection{Ablation Studies}\label{sec:ablations}
In this section we consider how the choice of MMD kernel, timestep scheduling and sampling process affect the performance of the proposed method. In all the experiments, unless otherwise specified, we use the DDIM sampling procedure and the Inception-V3 feature space.
First, we ablate the choice of the kernel for the MMD loss by comparing three different kernels: the linear kernel $k^\text{lin}(\vx, \vy) = \vx^\top\vy$, the polynomial cubic kernel $k^\text{cub}(\vx, \vy) = \left(\frac{1}{d}\vx^\top \vy+1\right)^3$ \cite{binkowski2018demystifying} and the Gaussian RBF kernel $k^\text{rbf}(\vx,\vy) = \exp\left( - \frac{1}{2 \sigma^2} \lVert x - y \rVert^2 \right)$. Table \ref{kernel-abl} reports the results for different kernels in terms of FID, showing a marginal preference for the cubic kernel and overall robustness of MMD-DDM to kernel choice.

\begin{table}%[h!]
    \caption{Ablation study for the kernel choice - CIFAR-10.}
    \vspace{10pt}
    \label{kernel-abl}
    \begin{center}
        \begin{tabular}{l c c c}
        \hline
        Kernel &  $\vert \mathcal{T} \vert=5$ & $\vert \mathcal{T} \vert=10$ & $\vert \mathcal{T} \vert=20$ \\
        \hline\hline
        Linear & 5.61 & 4.69 & 4.06  \\
        Gaussian RBF & 5.89 & 3.88 & 3.61 \\
        Cubic & \textbf{5.48}   & \textbf{3.80}   & \textbf{3.55} \\        
        \hline
    \end{tabular}
    \end{center}
\end{table}

Next, we ablate the influence of timesteps selection in Table \ref{timesteps-abl}. We consider the two commonly-used alternatives to select $\mathcal{T}$: \textit{linear} $\tau_i=\floor{ci}$, and \textit{quadratic} $\tau_i = \floor{ci^2}$, where $c$ is selected to make $\tau_1 \approx T$. This experiment does not show a preference for either selection method. However, it is possible that other subset selection strategies such as grid search \cite{dockhorngenie} or learning the optimal timesteps \cite{watson2022learning} could further improve results. We remark that MMD-DDM is decoupled from the specific timesteps selection technique.

\begin{table}%[h!]
    \caption{Ablation study for the timestep schedule - CIFAR10.}
    \vspace{10pt}
    \label{timesteps-abl}
    \begin{center}
        \begin{tabular}{l c c c}
        \hline
        Selection Method &  $\vert \mathcal{T} \vert=5$ & $\vert \mathcal{T} \vert=10$ & $\vert \mathcal{T} \vert=20$\\
        \hline\hline
        Linear & 5.48 & 3.80 & \textbf{3.55}  \\
        Quadratic & \textbf{5.19}  & 3.80   & 3.67  \\       
        \hline
    \end{tabular}
    \end{center}
\end{table}

Finally, we also test MMD-DDM with the DDPM sampling procedure, instead of DDIM. Results are reported in Table \ref{sampling-abl}. As expected, the DDIM sampling procedure is more powerful and produces better results with a low number of timesteps. However, we notice that MMD-DDM produces significant improvements even when applied to DDPM.

\begin{table}%[h!]
    \caption{Ablation study for the sampling procedure - CIFAR10.}
    \vspace{10pt}
    \label{sampling-abl}
    \begin{center}
        \begin{tabular}{l c c c}
        \hline
        Sampling & $\vert \mathcal{T} \vert=5$ & $\vert \mathcal{T} \vert=10$ & $\vert \mathcal{T} \vert=20$\\
        \hline\hline
        DDPM \cite{ho2020denoising}& 76.3 & 42.1 & 25.9 \\
        DDIM~\cite{song2020denoising} & 32.7 & 13.6 & 7.50 \\
        \hline
        DDPM+MMD-DDM & 6.65 & 5.19 & 4.48  \\
        DDIM+MMD-DDM& \textbf{5.48}   & \textbf{3.80}   & \textbf{3.55} \\      
        \hline
        
    \end{tabular}
    \end{center}
\end{table}

%----------------------------------------------------------------------------------------
\section{Conclusions and Discussion}
This paper addressed the problem of inference speed of DDMs. We showed that finetuning a DDM with a constraint on the number of timesteps using the MMD loss provides substantial improvements in visual quality. The limited computational complexity of the finetuning procedure offers a way to quickly obtain an improved tradeoff between inference speed and visual quality for a wide range of DDM designs. A limitation of the current technique lies in the memory requirements when the finetuning needs to be performed over a larger number of timesteps, although gradient checkpoint partially addresses this issue in most practical settings. Furthermore, coupling MMD-DDM with more advanced timestep selection and optimization techniques, possibly via joint optimization, could represent an interesting avenue to further improve speed-quality tradeoffs. Integration with conditional DDMs could also represent a direction for future work.

\paragraph{Broader Impact}

The goal of our method is accelerate synthesis in DDMs, which can make them more attractive methods for time-critical applications, and also reduce DDMs’ environmental footprint by decreasing the computational load during inference. However, it is well known (e.g \cite{vaccari}) that generative models can have unethical uses and potential bias depending on the context and datasets of the specific use cases. Therefore, practitioners should apply caution and mitigate impacts when using generative modeling for various applications.


\section{Appendix}

\subsection{Image Generation Results}
\begin{figure*}[t!]
\vspace{-0.4cm}
\begin{center}
{
\begin{tabular}{c@{\hspace{.15cm}}c@{\hspace{.25cm}}c@{\hspace{.25cm}}c}
      & DDIM & MMD-DDM(Inception-V3) & MMD-DDM(CLIP)\\
    \raisebox{1.5cm}{\rotatebox{90}{5 steps}}&
     \makebox{\includegraphics[width=.29\linewidth]{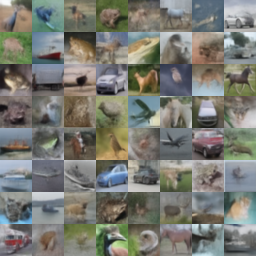}} & 
     \makebox{\includegraphics[width=.29\linewidth]{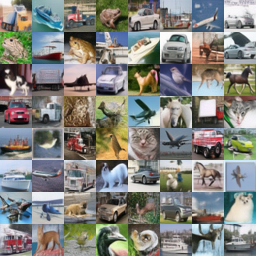}} & 
     \makebox{\includegraphics[width=.29\linewidth]{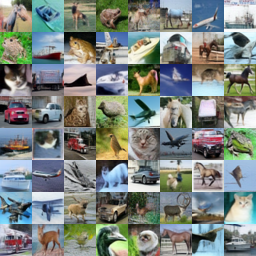}}\\ 
     \raisebox{1.5cm}{\rotatebox{90}{10 steps}}&
     \makebox{\includegraphics[width=.29\linewidth]{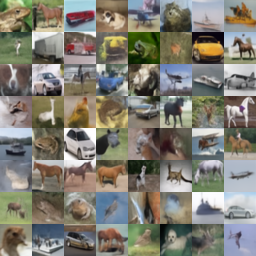}} & 
     \makebox{\includegraphics[width=.29\linewidth]{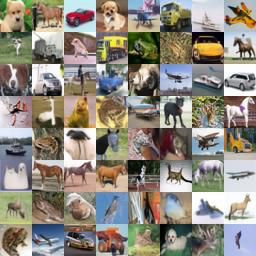}} & 
     \makebox{\includegraphics[width=.29\linewidth]{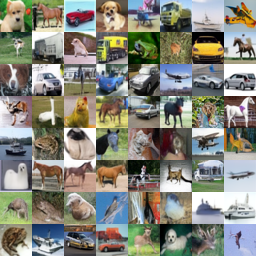}}\\
     \raisebox{1.5cm}{\rotatebox{90}{15 steps}}&
     \makebox{\includegraphics[width=.29\linewidth]{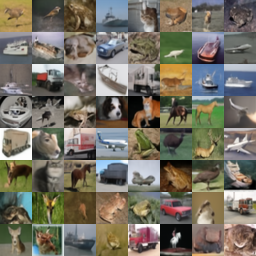}} & 
     \makebox{\includegraphics[width=.29\linewidth]{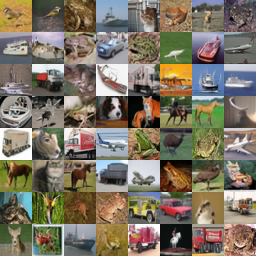}} & 
     \makebox{\includegraphics[width=.29\linewidth]{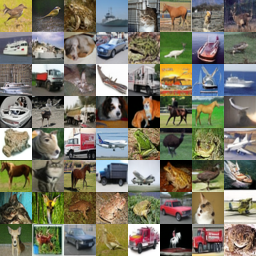}}\\ 
     \raisebox{1.5cm}{\rotatebox{90}{20 steps}}&
     \makebox{\includegraphics[width=.29\linewidth]{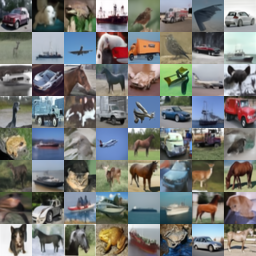}} & 
     \makebox{\includegraphics[width=.29\linewidth]{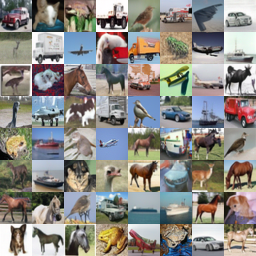}} & 
     \makebox{\includegraphics[width=.29\linewidth]{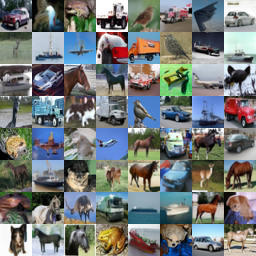}}
     
\end{tabular}
}
\end{center}

\caption{Generated samples for CIFAR-10. The samples are obtained using 5, 10, 15 and 20 timesteps with the DDIM sampling procedure. Results from Standard DDIM (left), the same model finetuned using Inception-V3 features (center) and CLIP features (right). 
\label{fig:cifar}}
\vspace{-0.1cm}
\end{figure*}

\begin{figure*}[t!]
\vspace{-0.4cm}
\begin{center}
{
\begin{tabular}{c@{\hspace{.15cm}}c@{\hspace{.25cm}}c@{\hspace{.25cm}}c}
      & DDIM & MMD-DDM (Inception-V3) & MMD-DDM (CLIP)\\
    \raisebox{1.5cm}{\rotatebox{90}{5 steps}}&
     \makebox{\includegraphics[width=.29\linewidth]{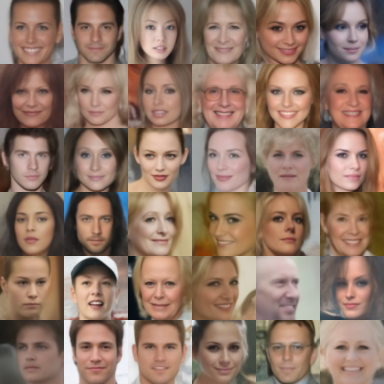}} & 
     \makebox{\includegraphics[width=.29\linewidth]{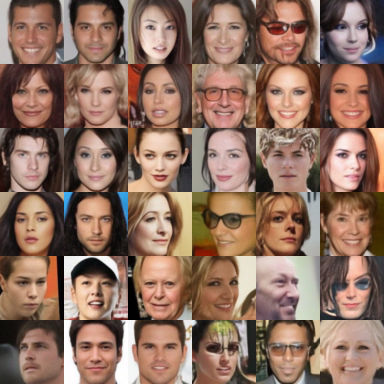}} & 
     \makebox{\includegraphics[width=.29\linewidth]{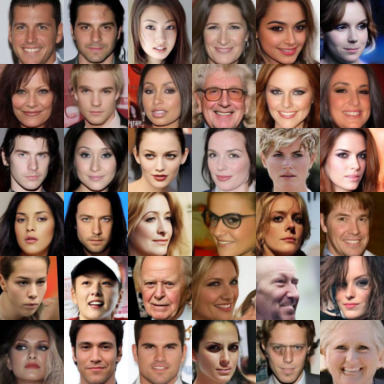}}\\ 
     \raisebox{1.5cm}{\rotatebox{90}{10 steps}}&
     \makebox{\includegraphics[width=.29\linewidth]{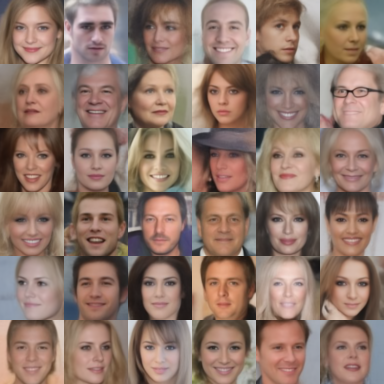}} & 
     \makebox{\includegraphics[width=.29\linewidth]{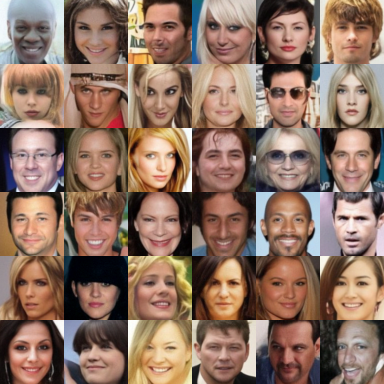}} & 
     \makebox{\includegraphics[width=.29\linewidth]{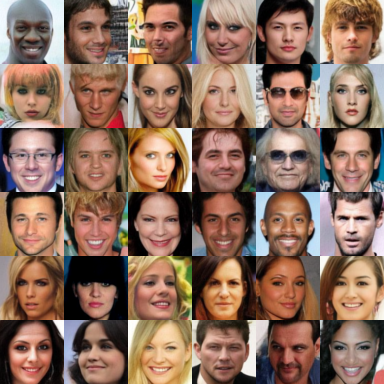}}\\
     \raisebox{1.5cm}{\rotatebox{90}{15 steps}}&
     \makebox{\includegraphics[width=.29\linewidth]{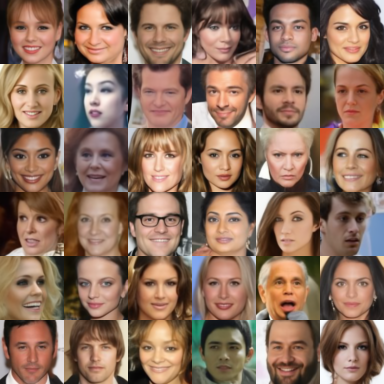}} & 
     \makebox{\includegraphics[width=.29\linewidth]{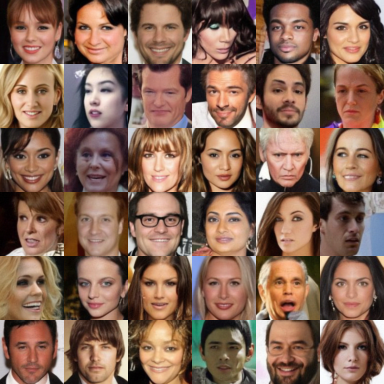}} & 
     \makebox{\includegraphics[width=.29\linewidth]{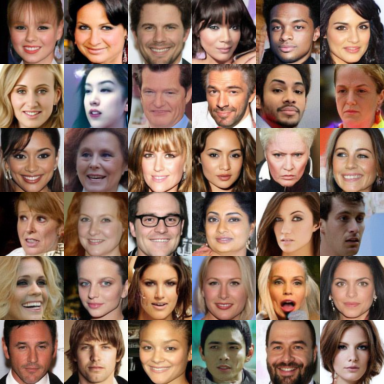}}\\ 
     \raisebox{1.5cm}{\rotatebox{90}{20 steps}}&
     \makebox{\includegraphics[width=.29\linewidth]{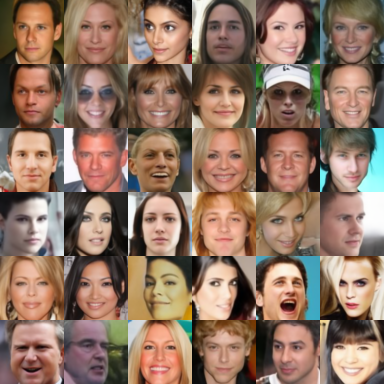}} & 
     \makebox{\includegraphics[width=.29\linewidth]{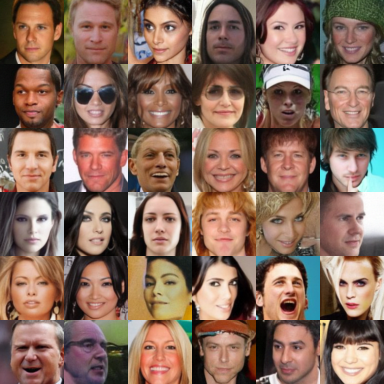}} & 
     \makebox{\includegraphics[width=.29\linewidth]{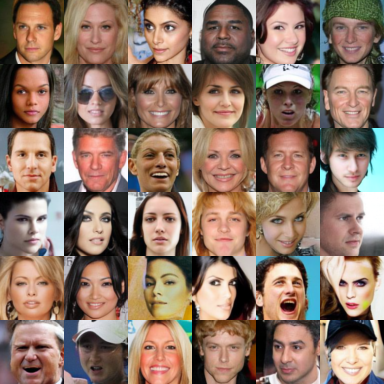}}
     
\end{tabular}
}
\end{center}

\caption{Generated samples for CelebA. The samples are obtained using 5, 10, 15 and 20 timesteps with the DDIM sampling procedure. Results from Standard DDIM (left), the same model finetuned using Inception-V3 features (center) and CLIP features (right). 
\label{fig:celeb}}
\vspace{-0.1cm}
\end{figure*}

\begin{figure*}[t!]
\vspace{-0.4cm}
\begin{center}
{
\begin{tabular}{c@{\hspace{.15cm}}c@{\hspace{.25cm}}c@{\hspace{.25cm}}c}
      & DDIM & MMD-DDM (Inception-V3) & MMD-DDM (CLIP)\\
    \raisebox{1.5cm}{\rotatebox{90}{5 steps}}&
     \makebox{\includegraphics[width=.29\linewidth]{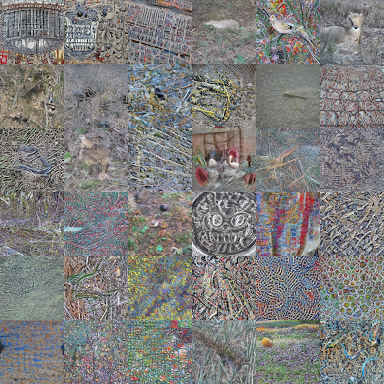}} & 
     \makebox{\includegraphics[width=.29\linewidth]{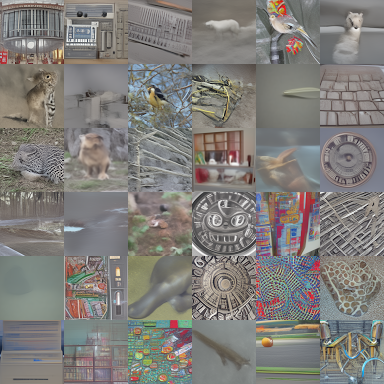}} & 
     \makebox{\includegraphics[width=.29\linewidth]{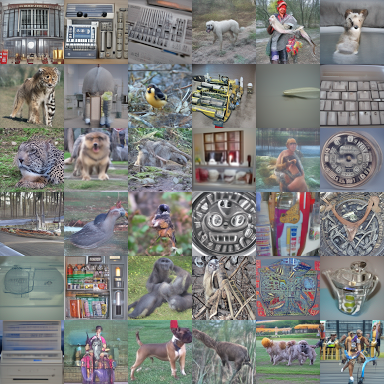}}\\ 
     \raisebox{1.5cm}{\rotatebox{90}{10 steps}}&
     \makebox{\includegraphics[width=.29\linewidth]{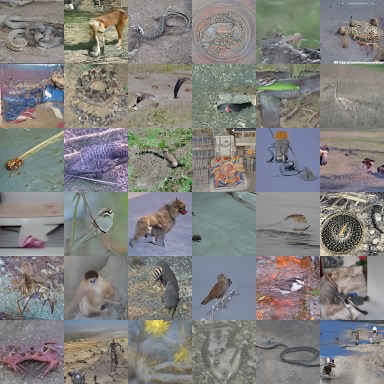}} & 
     \makebox{\includegraphics[width=.29\linewidth]{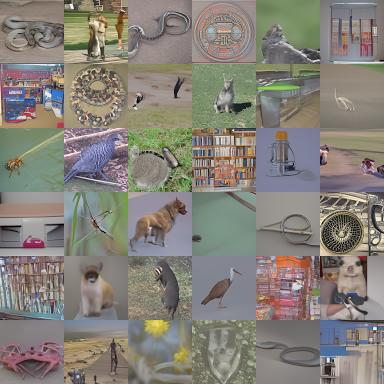}} & 
     \makebox{\includegraphics[width=.29\linewidth]{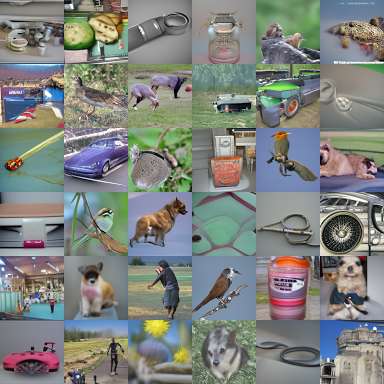}}\\
     \raisebox{1.5cm}{\rotatebox{90}{20 steps}}&
     \makebox{\includegraphics[width=.29\linewidth]{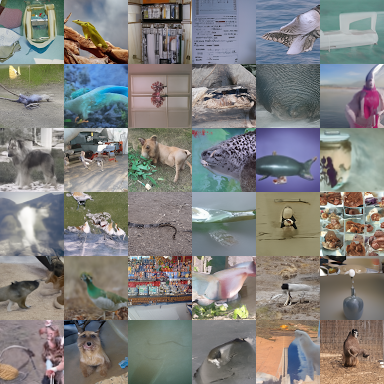}} & 
     \makebox{\includegraphics[width=.29\linewidth]{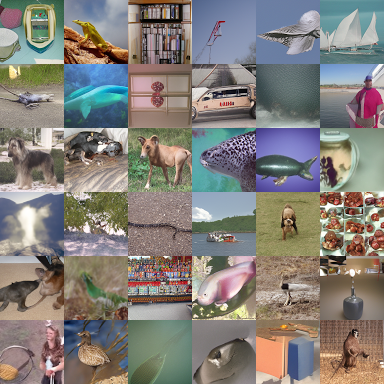}} & 
     \makebox{\includegraphics[width=.29\linewidth]{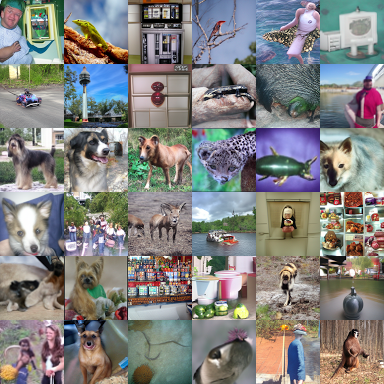}}
     
\end{tabular}
}
\end{center}

\caption{Generated samples for ImageNet. The samples are obtained using 5, 10 and 20 timesteps with the DDIM sampling procedure. Results from Standard DDIM (left), the same model finetuned using Inception-V3 features (center) and CLIP features (right). 
\label{fig:imagenet}}
\vspace{-0.1cm}
\end{figure*}

\begin{figure*}[t!]
\vspace{-0.4cm}
\begin{center}
{
\begin{tabular}{c@{\hspace{.15cm}}c@{\hspace{.25cm}}c@{\hspace{.25cm}}c}
      & DDIM & MMD-DDM (Inception-V3) & MMD-DDM (CLIP)\\
    \raisebox{1.5cm}{\rotatebox{90}{5 steps}}&
     \makebox{\includegraphics[width=.29\linewidth]{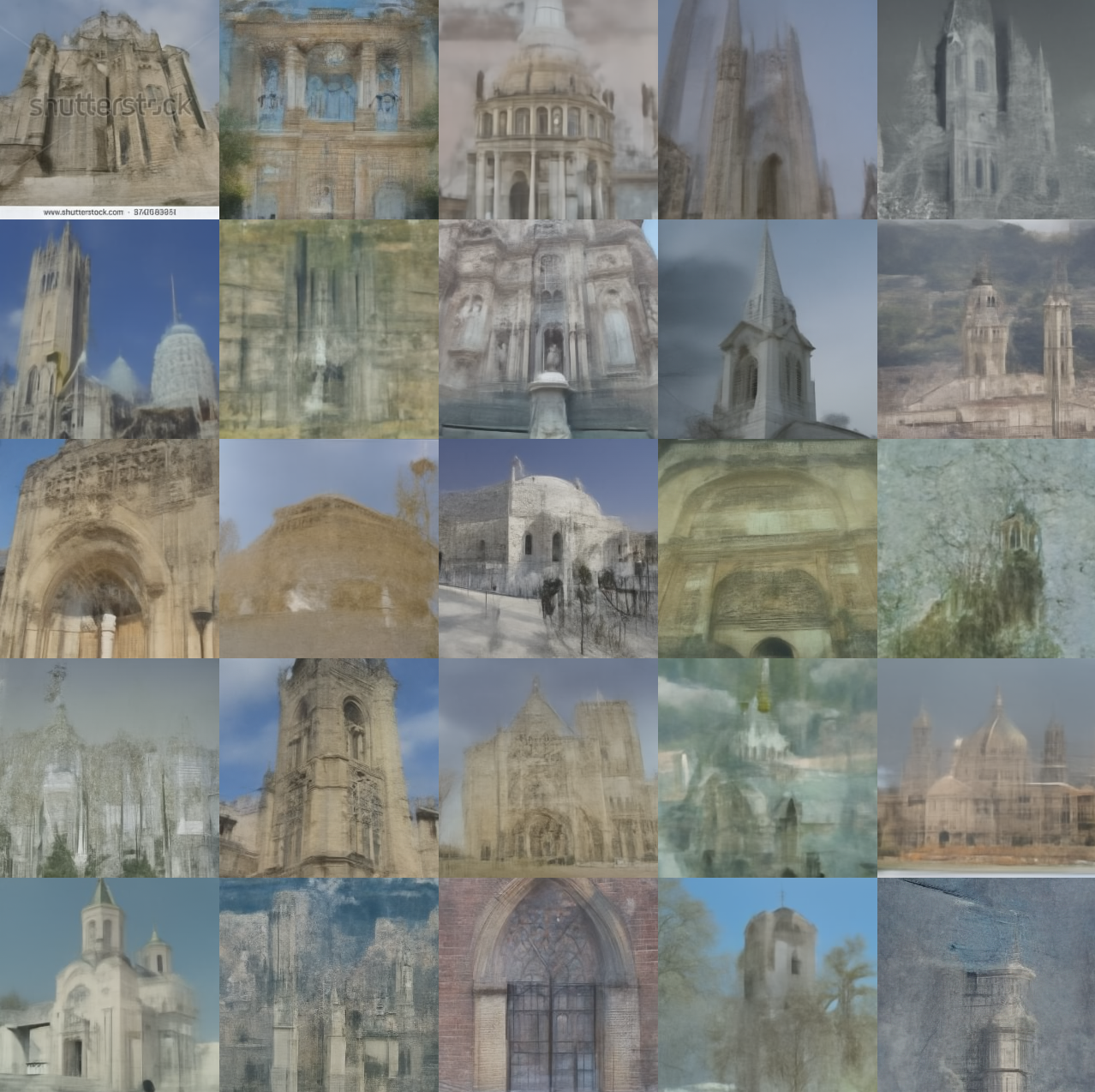}} & 
     \makebox{\includegraphics[width=.29\linewidth]{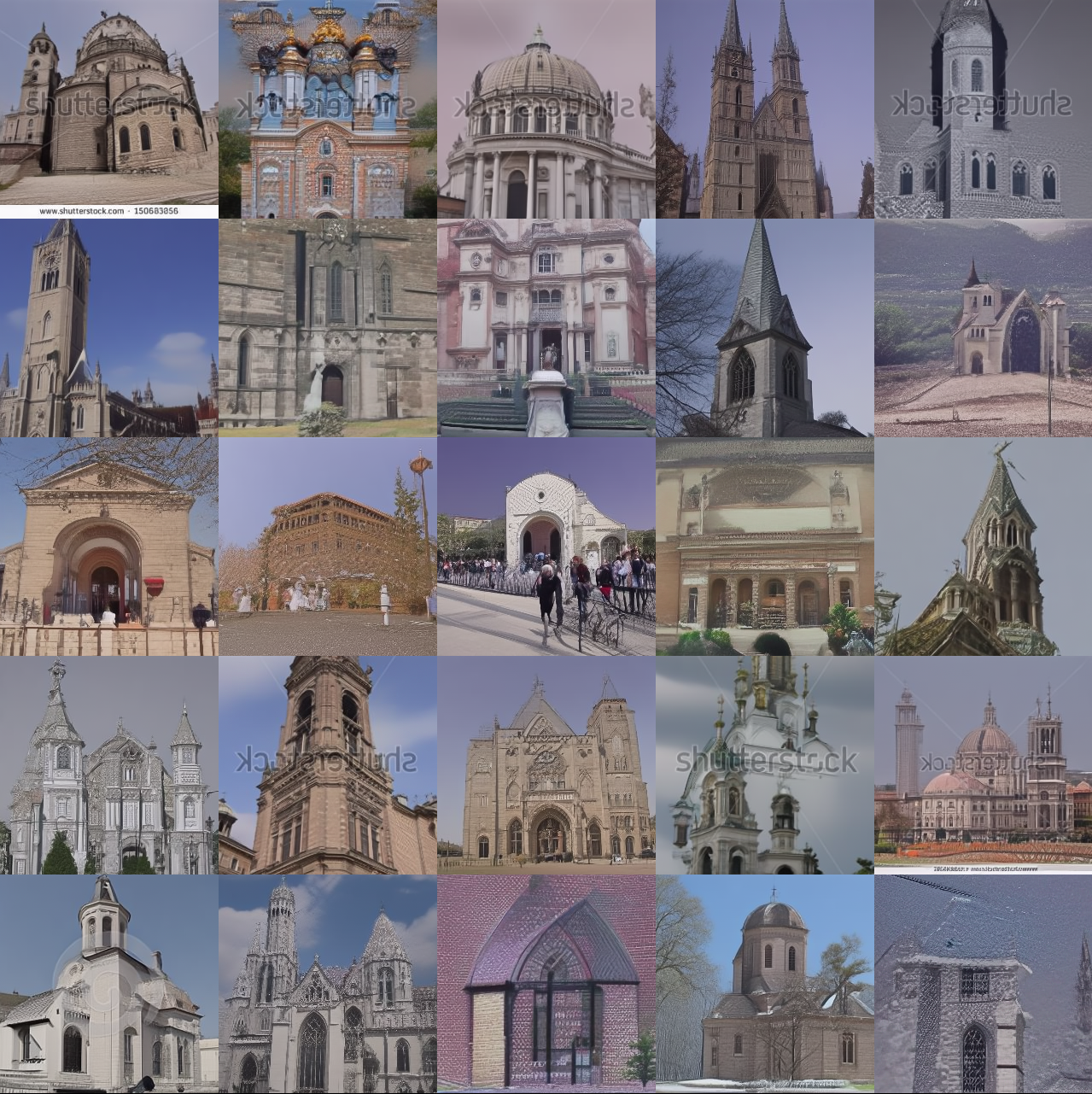}} & 
     \makebox{\includegraphics[width=.29\linewidth]{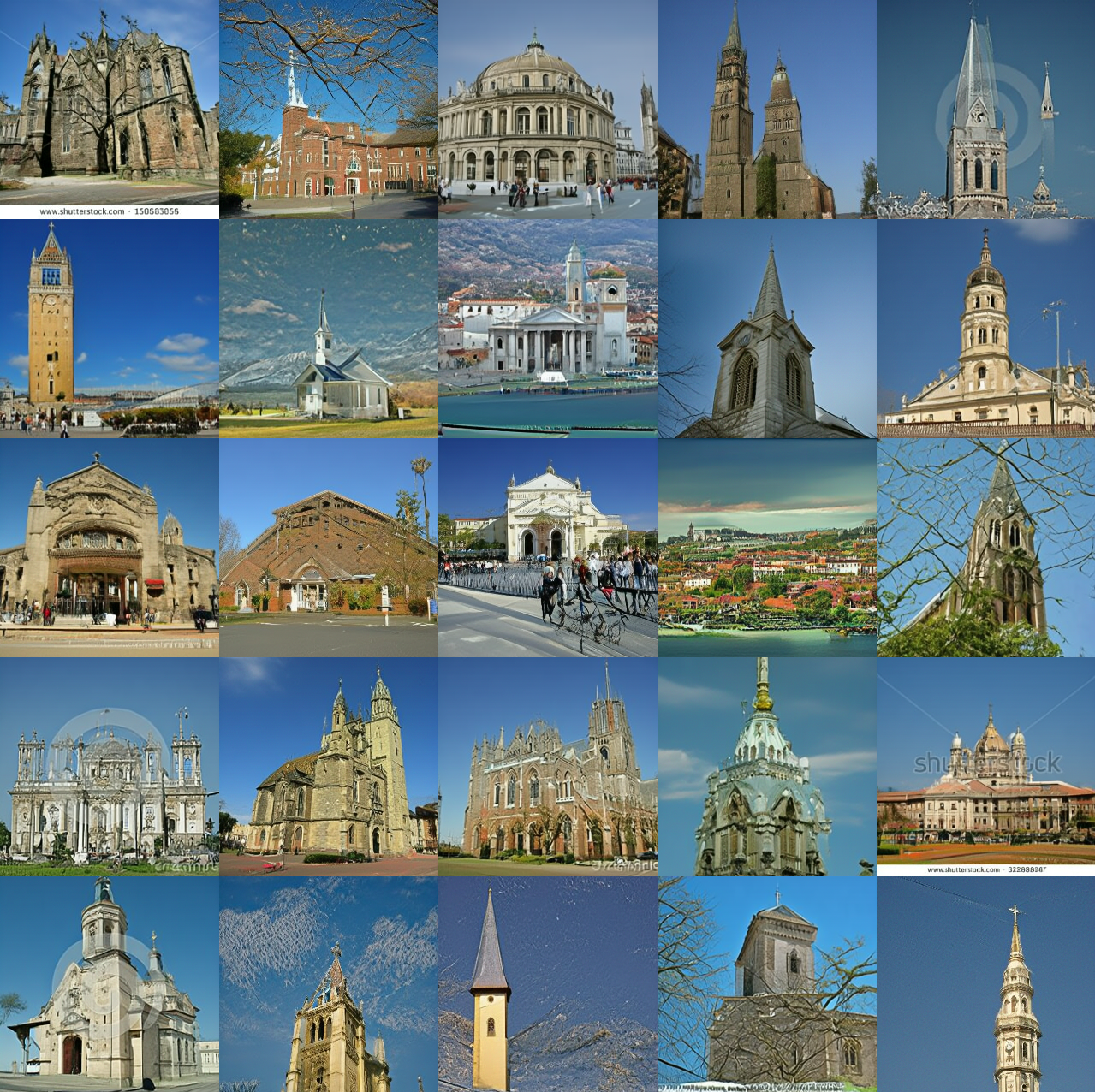}}\\ 
     \raisebox{1.5cm}{\rotatebox{90}{10 steps}}&
     \makebox{\includegraphics[width=.29\linewidth]{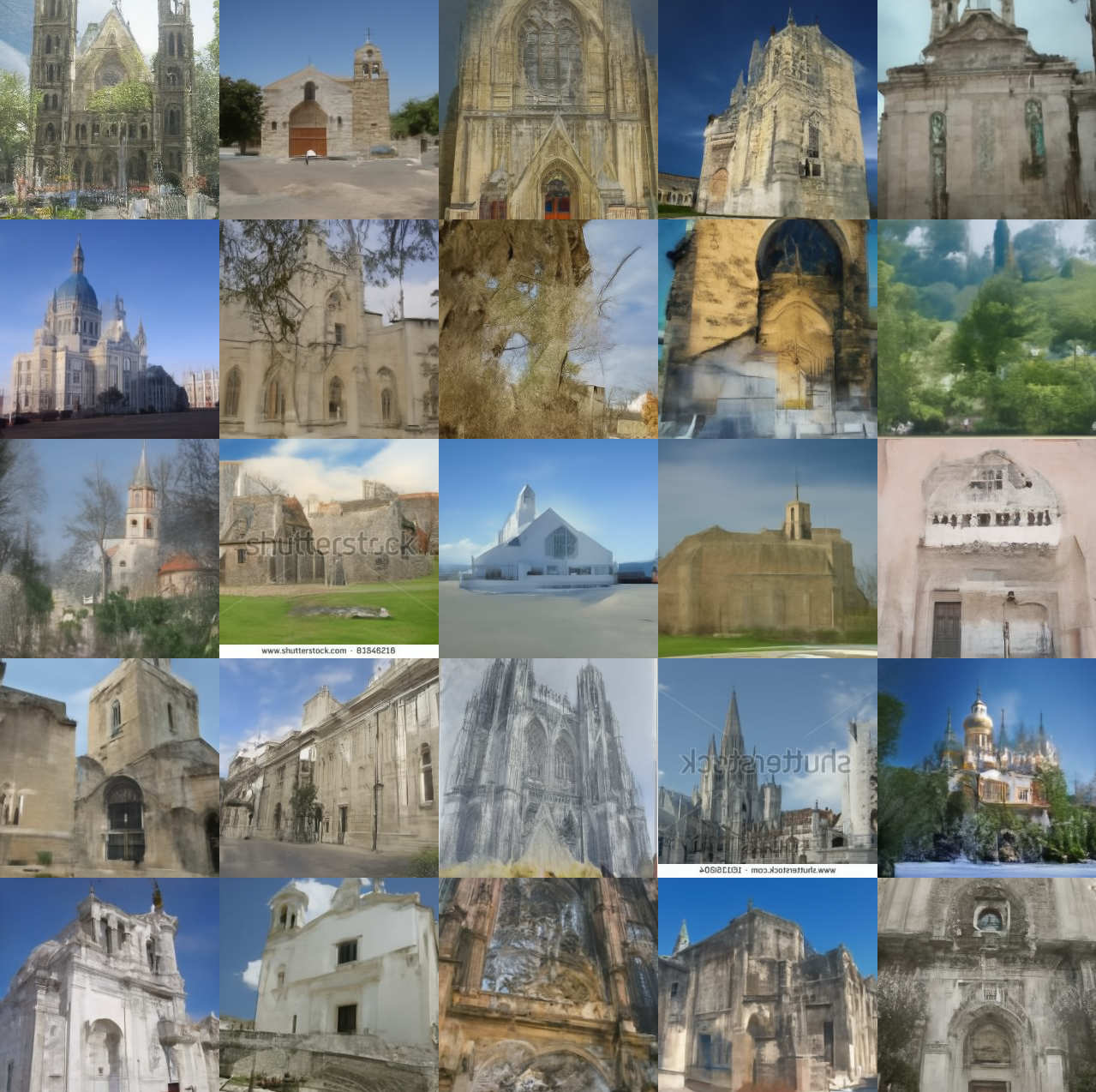}} & 
     \makebox{\includegraphics[width=.29\linewidth]{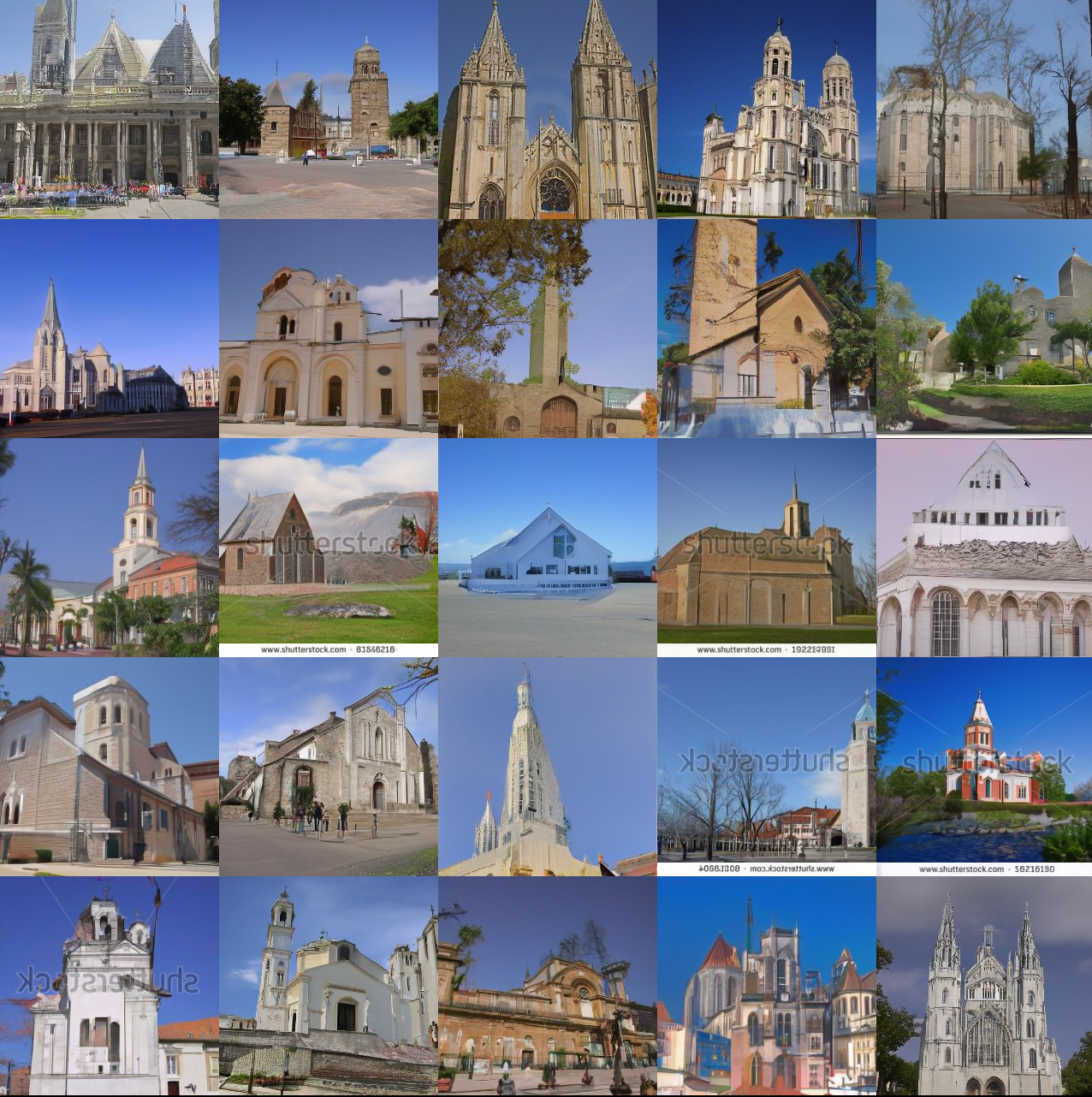}} & 
     \makebox{\includegraphics[width=.29\linewidth]{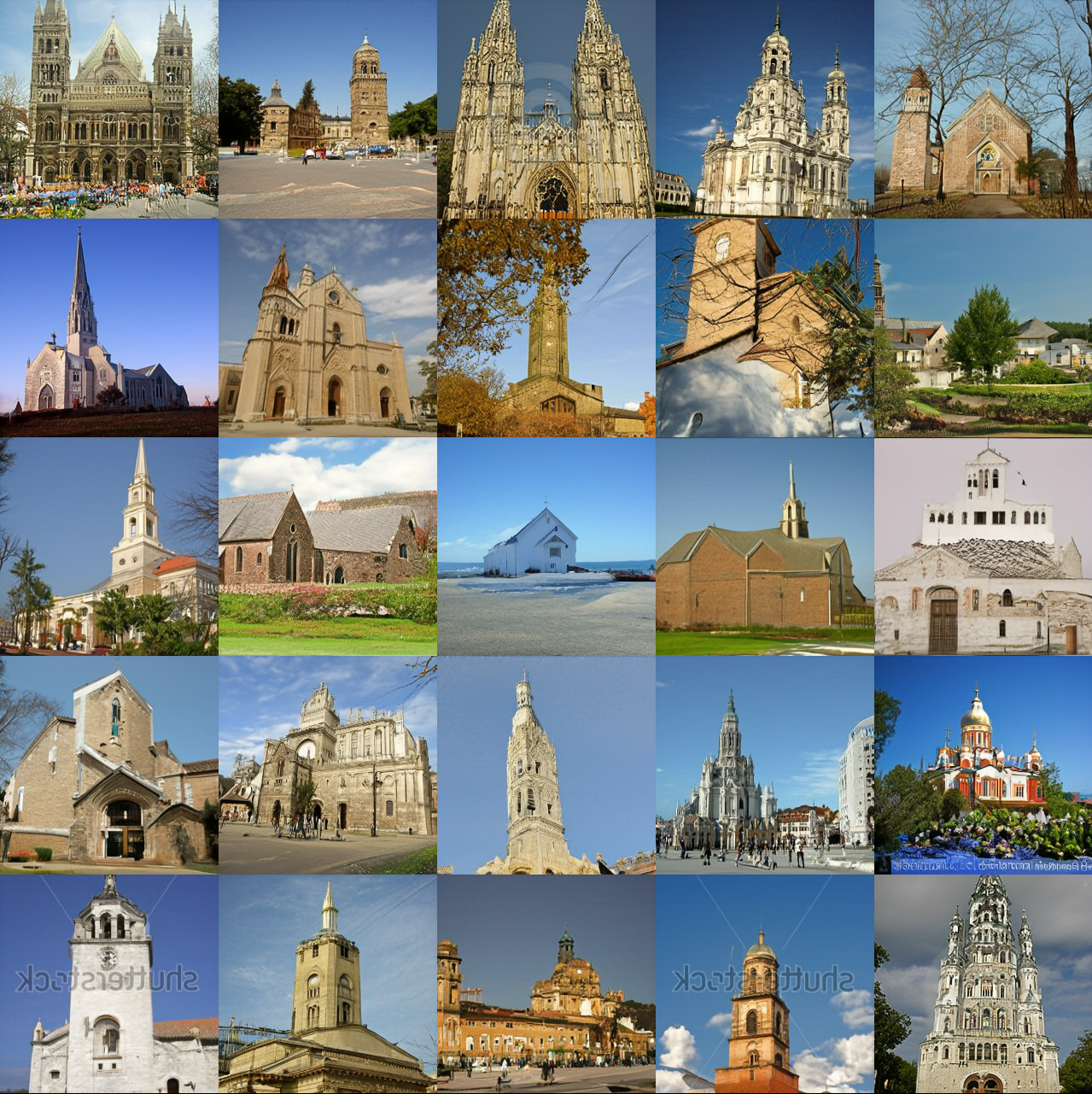}}\\
     \raisebox{1.5cm}{\rotatebox{90}{20 steps}}&
     \makebox{\includegraphics[width=.29\linewidth]{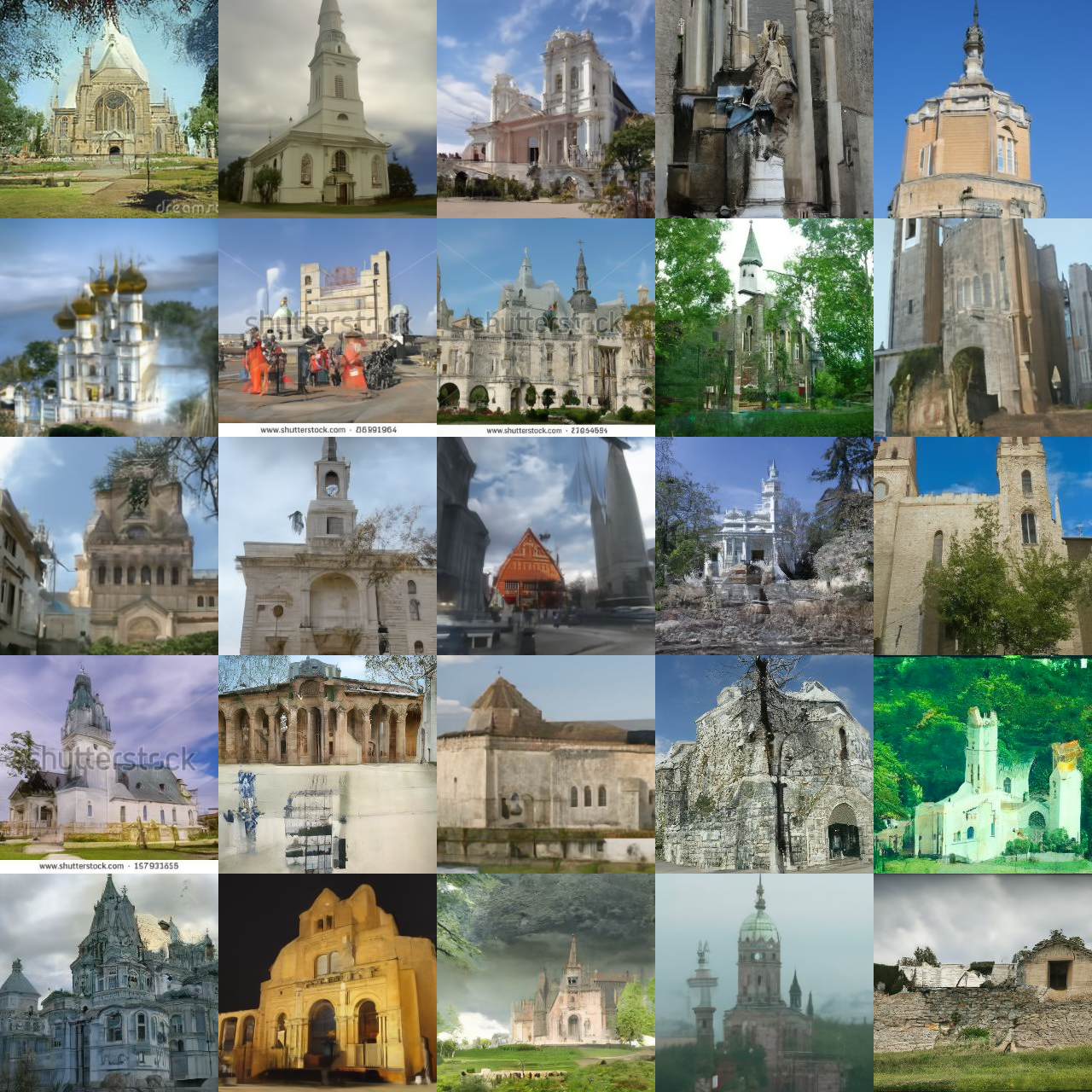}} & 
     \makebox{\includegraphics[width=.29\linewidth]{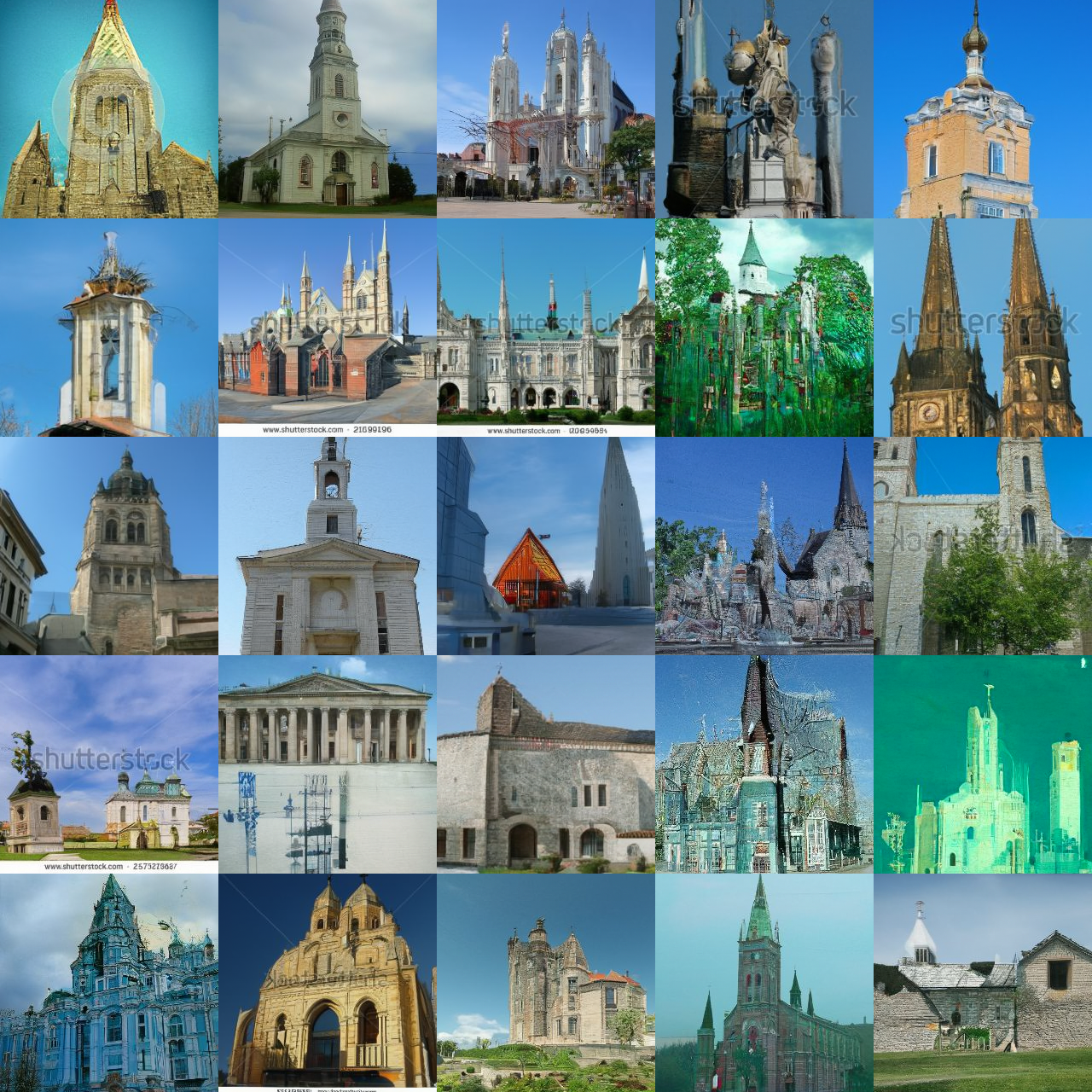}} & 
     \makebox{\includegraphics[width=.29\linewidth]{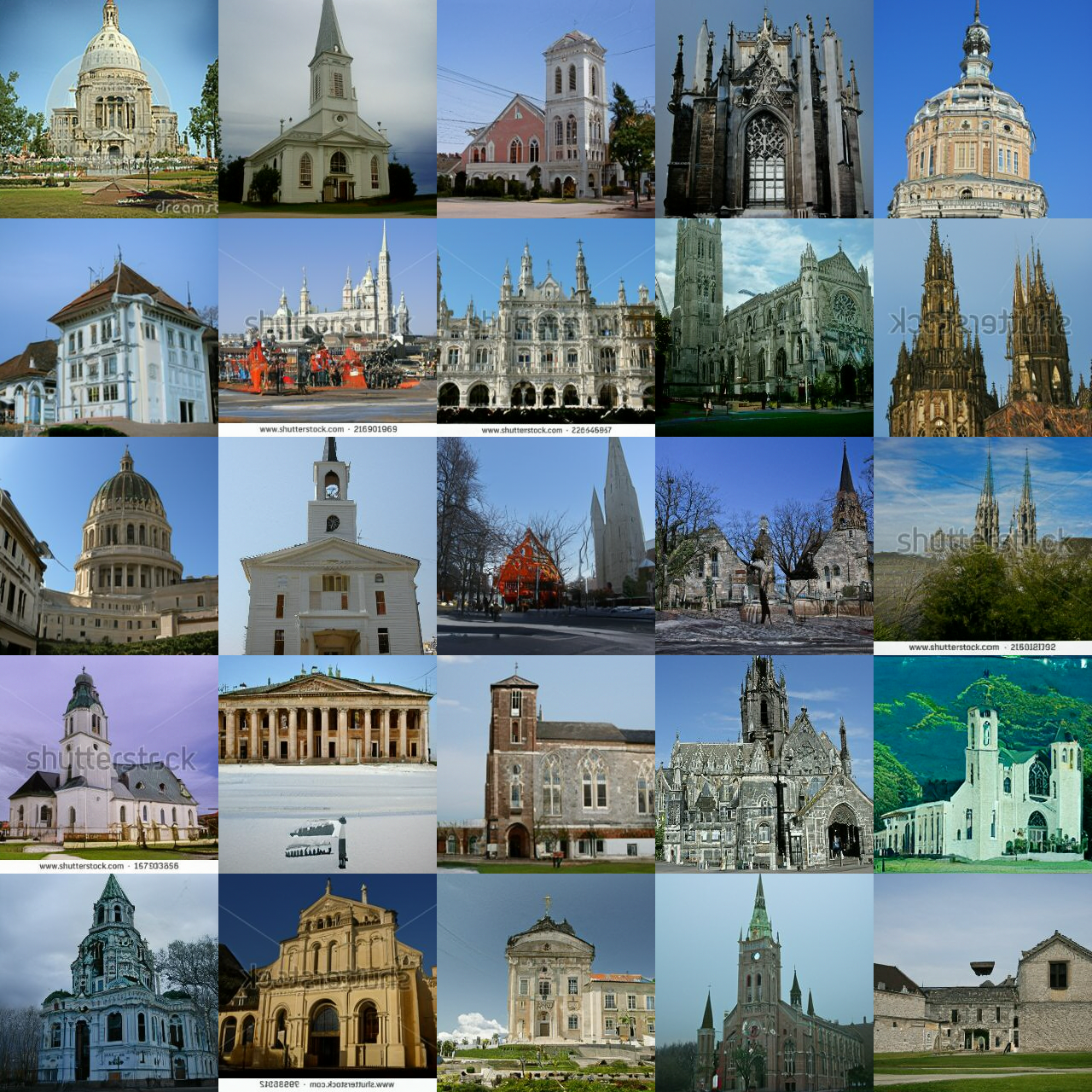}}
     
\end{tabular}
}
\end{center}

\caption{Generated samples for LSUN-Church Outdoor. The samples are obtained using 5, 10 and 20 timesteps with the DDIM sampling procedure. Results from Standard DDIM (left), the same model finetuned using Inception-V3 features (center) and CLIP features (right). 
\label{fig:lsun}}
\vspace{-0.1cm}
\end{figure*}

We report visualizations of generated images on different datasets with several timesteps. Figure \ref{fig:cifar} reports the results for CIFAR, Figure \ref{fig:celeb} while Figs. \ref{fig:imagenet} and \ref{fig:lsun} for ImageNet and LSUN-Church respectively.

\vspace{-0.07in}
\subsection{Additional Evaluation Metrics}

We report additional evaluation metrics computed on CIFAR-10. The Inception Score \cite{salimans2016improved}, the Spatial FID \cite{nash2021generating} , i.e. the FID evaluated using the first 7 channels from the intermediate \textit{mixed\_6/conv} feature maps, and the Precision and Recall metrics \cite{sajjadi2018assessing}. In particular Recall is used to evaluate the diversity of generated samples. We denote an overall improvement on all the metrics for both our proposed solutions. The results are showed in Table \ref{cifar-res-extended}. Interestingly our finetuning increases the diversity of the generated samples leading to higher recall scores. \footnote{The values are obtained using OpenAI evaluator that can be found at: https://github.com/openai/guided-diffusion/tree/main/evaluations}

\begin{table*}[t]
    \caption{Unconditional CIFAR-10 generative performance over different metrics.}
    \vspace{10pt}
    \label{cifar-res-extended}
    \begin{center}
        \begin{tabular}{l c c c c}
        \hline
        Method &  $\vert \mathcal{T} \vert=5$ & $\vert \mathcal{T} \vert=10$ & $\vert \mathcal{T} \vert=15$ & $\vert \mathcal{T} \vert=20$  \\%& NFEs=25 \\ %
        \hline\hline
        \textit{Inception Score} & & & & \\
        \hline
        DDIM~\cite{song2020denoising} & 6.95 & 8.17 & 8.43 & 8.54 \\
        MMD-DDM (Inception-V3) &  \textbf{9.39}  &  \textbf{9.91}  & \textbf{9.85} &  \textbf{9.94}\\
        MMD-DDM (CLIP) & 9.10  & 9.38 & 9.52 & 9.19 \\
        \hline
        \textit{sFID} & & & & \\
        \hline
        DDIM~\cite{song2020denoising} & 22.2 & 10.73 & 9.35 & 7.36 \\
        MMD-DDM (Inception-V3) &  \textbf{7.20}  &  6.09  &  \textbf{5.75}  &  5.46 \\
        MMD-DDM (CLIP) & 8.25 & \textbf{6.07} & 6.04 & \textbf{4.90} \\
        \hline
        \textit{Precision} & & & & \\
        \hline
         DDIM~\cite{song2020denoising} & 0.51 & 0.60 & 0.61 & 0.63 \\
        MMD-DDM (Inception-V3) & 0.63   &  0.65  & 0.65 &  0.67\\
        MMD-DDM (CLIP) & \textbf{0.65}  & \textbf{0.67} &  \textbf{0.67} & \textbf{0.68} \\
        \hline
        \textit{Recall} & & & & \\
        \hline
        DDIM~\cite{song2020denoising} & 0.31 & 0.46 & 0.50 & 0.52 \\
        MMD-DDM (Inception-V3) & \textbf{0.56}  & \textbf{0.59} & \textbf{0.60} &  \textbf{0.58} \\
        MMD-DDM (CLIP) & 0.53  & 0.56 & 0.57 & 0.57 \\
        \hline

    \end{tabular}
    \end{center}
\end{table*}

\begin{figure*}
\vspace{-0.4cm}
\begin{center}
\includegraphics[width=0.8\linewidth]{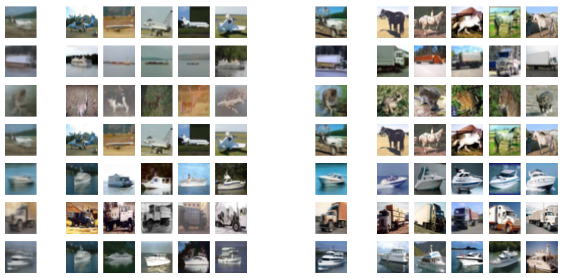}\\
\vspace{1.5cm}
\includegraphics[width=0.8\linewidth]{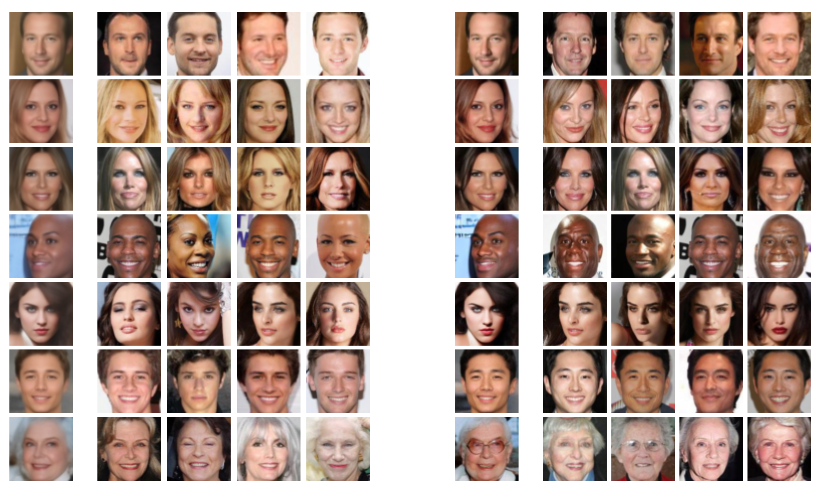}\\
\caption{Generated samples by the DDIM model (left) and the finetuned model (right) for CIFAR-10 (top) and CelebA (bottom). For each generated samples we visualize the top-k nearest neighbours.\label{fig:nn}}
\end{center}
\end{figure*}

\vspace{-0.07in}
\subsection{Nearest Neighbours Visualization}

We visualize the top-K nearest neighbors of several samples in the CLIP feature space for both the pre-trained model and the fine-tuned model with the Inception feature space as explained in Section 4.4. Results are reported in Figure \ref{fig:nn}.

\begin{figure*}[t!]
\vspace{-0.4cm}
\begin{center}
\includegraphics[width=0.8\linewidth]{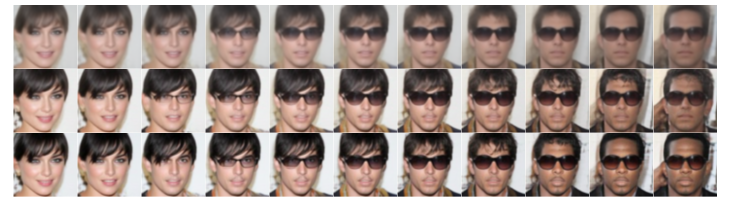}\\
\vspace{1in}
\caption{Generated interpolations by the DDIM model (first row) and the Inception finetuned model (second row) and the CLIP finetuned model (third row) for CelebA. The samples are generated using 5 timesteps.\label{fig:celeb_int}}
\end{center}
\end{figure*}

\begin{figure*}
\vspace{-0.4cm}
\begin{center}
\includegraphics[width=0.8\linewidth]{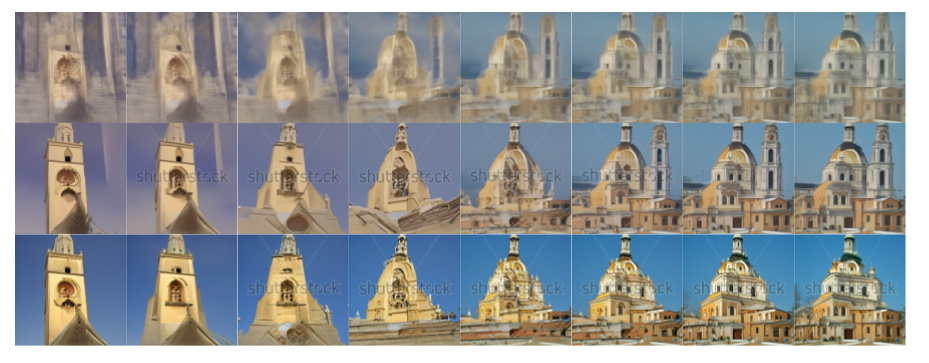}\\
\caption{Generated interpolations by the DDIM model (first row) and the Inception finetuned model (second row) and the CLIP finetuned model (third row) for LSUN-Church The samples are generated using 5 timesteps.\label{fig:lsun_int}}
\end{center}
\end{figure*}

\vspace{-0.07in}
\subsection{Latent Space Interpolation}
We create interpolations (Figs. \ref{fig:celeb_int} \ref{fig:lsun_int}) on a line by selecting two random $x_T$ values from a standard Gaussian distribution, using them in a spherical linear interpolation method  and then applying the DDIM sampling \cite{ho2020denoising} to generate $x_0$ samples. 
\begin{align}
    \vx_T^{(\alpha)} =  \frac{\sin((1 - \alpha) \theta)}{\sin(\theta)} \vx_{T}^{(0)} + \frac{\sin(\alpha \theta)}{\sin(\theta)} \vx_{T}^{(1)}
\end{align}
where $\theta = \arccos\left(\frac{(\vx_T^{(0)})^\top \vx_T^{(1)}}{\norm{\vx_T^{(0)}} \norm{\vx_T^{(1)}}}\right)$.

\end{document}